\documentclass{article}

\PassOptionsToPackage{numbers}{natbib}
\usepackage[eandd,preprint]{neurips_2026}

\usepackage[utf8]{inputenc}
\usepackage[T1]{fontenc}
\usepackage{amsmath,amssymb}
\usepackage{booktabs}
\usepackage{hyperref}
\usepackage{url}
\usepackage{graphicx}
\usepackage{enumitem}
\usepackage{microtype}
\usepackage{float}

\title{PhAIL: A Real-Robot VLA Benchmark and Distributional Methodology}

\author{
  Sergey Arkhangelskiy \\
  Positronic Robotics \\
  \texttt{s@positronic.ro}
}

\date{}

\begin{document}

\maketitle

\begin{abstract}
Real-world evaluation of vision-language-action (VLA) policies still rests on binary success rate at a fixed timeout with $N \le 25$ rollouts per condition, almost always without confidence intervals or paired statistical comparison -- cohort sizes that struggle to resolve close comparisons reliably. We introduce \textbf{PhAIL} (Physical AI Leaderboard, \url{https://phail.ai}), an open real-robot benchmark on a Franka FR3 -- dataset, per-rollout artifacts, and end-to-end reference implementation -- of a distributional evaluation methodology: the time-to-success cumulative distribution function (CDF) as the evaluation primitive, with two separated jobs -- \emph{scoring} (Human-Relative Throughput, HRT -- a dimensionless scalar with bootstrap confidence intervals, anchored to same-fixture human teleoperation) and a \emph{significance test} (Kolmogorov--Smirnov, computed per-object and macro-averaged across objects). On four publicly-available VLAs, the macro-averaged KS test resolves two close comparisons (GR00T vs.\ ACT, OpenPI vs.\ ACT) at $N \le 30$ rollouts per (model, object) cell where binary-threshold metrics do not -- the closest pair (OpenPI vs.\ GR00T) remains unresolved within our budget. The best evaluated VLA is ${\sim}7\times$ slower per operation (RMST ratio) than the human reference.
\end{abstract}

\section{Introduction}\label{sec:intro}

How do you tell whether one robot policy is better than another in the real world? Most published evaluations of vision-language-action (VLA) policies answer this with binary success rate at a fixed timeout, $N \le 25$ rollouts per condition, no confidence intervals, no paired test -- a depth that is an order of magnitude below the budget the binary tests they implicitly use require for reliable ranking. Sampling noise can therefore obscure genuine ordering effects on cohorts the field can practically grow.

The standard fix the field has reached for -- adding a throughput metric (units per hour (UPH), cycle time, completion fraction) alongside binary success -- works in part but surfaces a deeper problem: when the two scalars rank policies in different orders, the choice between them is task-dependent and rarely disclosed in print. Recent methodology critiques argue that blinded same-session A/B with hundreds of rollouts is feasible~\cite{kress2024,lbm2026}, and the closest ML-side precedent for distributional comparison~\cite{agarwal2021} extends the conversation toward distributions. None of these recommendations ships a primitive plus a reference implementation that lets the field apply them on a shared station.

We respond by adopting the time-to-success CDF as the evaluation primitive. Let $T$ be the wall-clock time the policy needs to complete one operation; hard failures (items lost outside the workspace, items dropped on the table at episode end, safety stops) are absorbed as $T = \infty$ ghost events rather than censored tails. The resulting CDF $F(t) = P(T \le t)$ jointly carries reliability (the gap between its asymptote and $1$ equals the hard-failure rate) and throughput (the median of $T$, and $\text{UPH} \propto 1/E[T \mid T < \infty]$). Standard scalars -- success rate at $\tau$, UPH, mean time between failures/assists (MTBF/A), completion fraction, restricted mean survival time (RMST) -- are projections of $F$, and the CDF is strictly richer than any of them: when two CDFs cross, no single scalar suffices and reasonable scalars give opposite signs on the same data. We separate two methodological jobs that current practice runs together: \emph{scoring} (a scalar with bootstrap confidence intervals (CIs) -- Human-Relative Throughput, HRT, anchored to a same-fixture human reference) and a \emph{significance test} (macro-averaged KS computed per-object on the time-to-success CDFs). PhAIL is the open real-world inference benchmark that instantiates this methodology on a Franka FR3 across four objects.

Empirically, on four publicly-available VLAs -- OpenPI $\pi_{0.5}$~\cite{pi05}, GR00T N1.6~\cite{groot_n1}, ACT~\cite{act}, SmolVLA~\cite{smolvla} -- the best is roughly $7\times$ slower per operation (RMST ratio) than the human reference, the top two are statistically indistinguishable on every metric tested, and SmolVLA is clearly worst. The choice of \emph{test}, not just the choice of $N$, determines what is resolvable: macro-averaged KS detects the GR00T-vs-ACT difference at $N{=}25$ and the OpenPI-vs-ACT difference at $N{=}30$ rollouts per (model, object) cell, while a stratified-McNemar binary baseline on a 5\,pp paired difference needs 600--1500 paired rollouts \emph{per cell}~\cite{mcnemar1947,connor1987} -- roughly $30\times$ more rollouts at the unit both tests stratify on (this ratio reflects our chosen binary baseline and per-object CDF shapes; the methodology, not the multiplier, is what generalizes). The field's modal $N \le 25$ is therefore an order of magnitude or more below the per-cell binary-test budget at our effect size.

\paragraph{Contributions.}

\begin{figure}[!t]
  \centering
  \includegraphics[width=\linewidth]{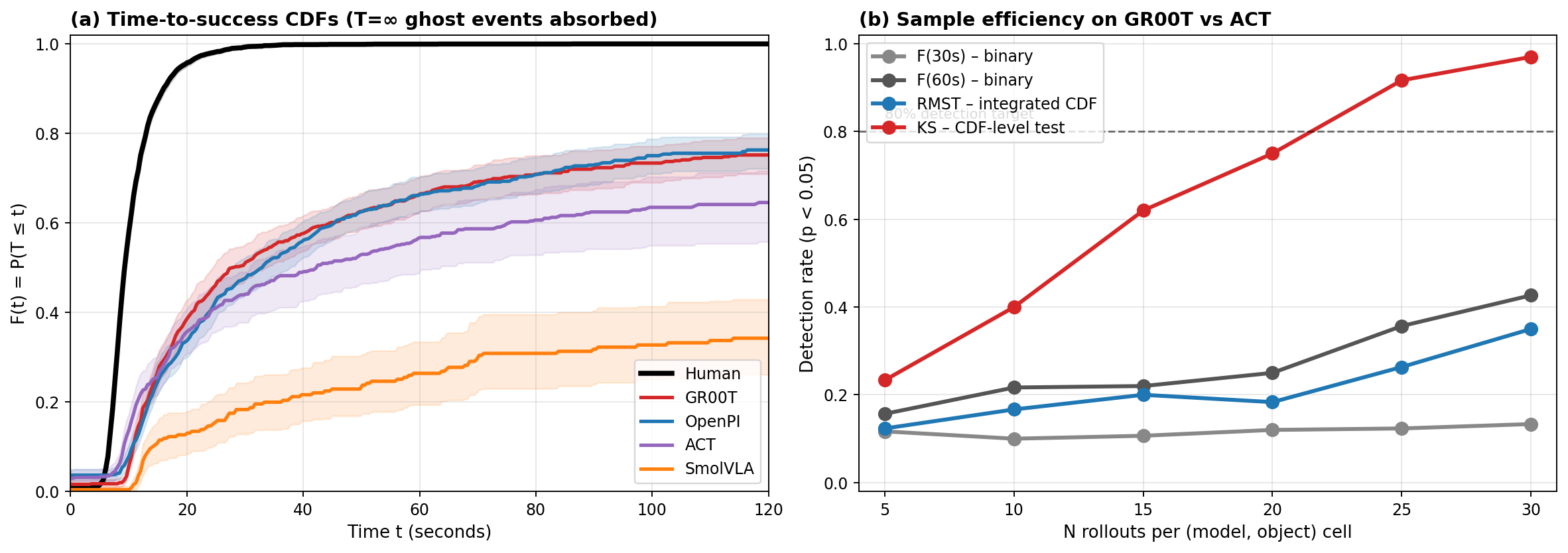}
  \caption{\textbf{(a)} Time-to-success CDFs are richer than any scalar: reliability and throughput on a single axis. The four VLAs all sit far below the human reference -- the best is $\sim$$7\times$ slower. Hard failures become the asymptote below $F=1$. \textbf{(b)} Choosing the right \emph{test}, not just running more trials, is what resolves close comparisons: macro-averaged Kolmogorov--Smirnov across per-object CDFs reaches 80\% detection on GR00T vs.\ ACT at $N{=}25$ rollouts per (model, object) cell, while binary success-rate-at-threshold ($F(30s)$, $F(60s)$) and integrated RMST fall well short of the target on every close pair across the entire $N \le 30$ range. The dominant practice ($F(\tau)$ at modal $N \le 25$) is underpowered for ranking close pairs at this depth.}
  \label{fig:hero}
\end{figure}

\begin{enumerate}[leftmargin=2em]
  \item \textbf{Time-to-success CDF as the evaluation primitive}, with hard failures absorbed as $T = \infty$ ghost events (Figure~\ref{fig:hero}a). Standard scalars are projections; we propose macro-averaged Human-Relative Throughput (HRT) as the dimensionless headline scalar grounded in a same-fixture human reference.
  \item \textbf{Same data, opposite rankings -- no single scalar suffices.} Principled aggregations of the same CDF disagree on top-1, so the choice of headline scalar is a disclosed methodological commitment.
  \item \textbf{For distinguishing close-pair policies, the macro-averaged KS test is roughly $30\times$ more sample-efficient per cell than stratified binary-threshold baselines} (Figure~\ref{fig:hero}b) -- an advantage rooted in test design rather than dataset specifics (\S\ref{sec:framework-test}). Scoring (HRT for ranking) and the significance test (KS for distinguishability) are separate jobs, and for the significance step the choice of test, not the choice of $N$, is what drives sample efficiency.
  \item \textbf{PhAIL: open real-world benchmark with auditable artifact.} Beyond the statistical methodology, four robotics-side contributions: (i) a real-robot dataset of $\sim$990 episodes (including 396 same-fixture human-teleoperation reference rollouts) on a Franka FR3, four trained objects; (ii) a same-fixture human-reference protocol that turns embodiment-specific timings into a dimensionless ratio; (iii) operationally-grounded failure semantics (recovery-impossible outcomes as $T{=}\infty$ ghost events, recoverable slow tails left censored); (iv) per-rollout synchronized video, telemetry, and event annotations auditable through a public run-explorer interface -- every annotation is reproducible from the raw rollout. Blinded same-session randomized rotation is the single protocol recommendation that does the most work (Appendix~\ref{app:spatial}).
\end{enumerate}

\paragraph{Scope.} The methodological contribution is the comparison primitive, the resolution procedure, the power budget, and the open framework that makes new claims auditable. \textbf{The methodology is task-agnostic} -- it applies to any operation with an unambiguous success event and a same-fixture human reference. PhAIL's current empirical validation is bin-to-bin pick-and-place across four objects (\S\ref{sec:phail}); subsequent releases will add insertion and small-part assembly, with the methodology unchanged across additions, and other natural fits include packing, navigation, and articulated manipulation. The per-model rankings are illustrative of what the methodology can resolve at our depth, not architectural pronouncements; we used each repository's default fine-tuning recipe (\S\ref{sec:phail-finetune}). PhAIL is open to submissions.

\section{Related Work}\label{sec:related}

\paragraph{Existing real-robot evaluation protocols.} RoboArena~\cite{roboarena} and RoboChallenge~\cite{robochallenge} use physical robots but compare via pairwise Elo or scalar success on standardized tasks; neither reports confidence intervals or paired tests on the rankings. Competition-style evaluations (RGMC, OCRTOC~\cite{ocrtoc}, NIST Assembly Task Boards~\cite{nist_atb}, the Real Robot Challenge~\cite{realrobotchallenge}, surveyed in~\cite{sun_competitions_2022}) standardize task scoring without statistical machinery; closest in spirit is the Digital Robot Judge~\cite{so2023}, which builds task-centric performance databases via electronic task boards but reports raw telemetry only. The most rigorous existing real-robot evaluation is the LBM examination~\cite{lbm2026}, which runs blinded same-session A/B at $N{=}50$--$200$ per condition with Bayesian posterior violins on success rate, paired Barnard's exact tests for binary outcomes, Welch's $t$-tests for a scalar task-completion score, and Bonferroni-corrected significance grouping -- a strong precedent for what rigorous real-robot evaluation looks like. PhAIL's contribution is orthogonal to LBM's: a richer primitive (time-to-success CDF with $T{=}\infty$ ghost events, jointly carrying reliability and throughput) and a same-fixture human-reference anchor that turns embodiment-specific timings into a dimensionless ratio. Simulated benchmarks (RoboCasa, CALVIN, ManiSkill, SIMPLER~\cite{robocasa,calvin,maniskill,simpler}) iterate fast but miss real-world perception, latency, and safety; they are complementary, not substitutes.

\paragraph{Methodology critiques.} Kress-Gazit et al.~\cite{kress2024} and STEP~\cite{snyder2025} argue current real-robot evaluation is statistically underpowered and recommend blinded same-session A/B at hundreds of rollouts; Agarwal et al.~\cite{agarwal2021} is the closest ML-side precedent for distributional comparison. We extend the distributional approach to time-to-event data with $T{=}\infty$ ghost events and ship a reference implementation on a shared station. A 13-paper survey of recent VLA evaluation practice (Appendix~\ref{app:survey}) confirms modal per-condition $N$ is 10--20 and none of the 13 standard-practice papers report confidence intervals or paired tests; LBM is the single recent counter-example.

\section{Time-to-Success CDF as Evaluation Primitive}\label{sec:framework}

The methodology has three components: the random variable and its CDF (\S\ref{sec:framework-T}), the scoring layer (scalars and visualizations derived from the CDF, \S\ref{sec:framework-scoring}), and the significance layer (distributional tests on the CDF, \S\ref{sec:framework-test}). The strict separation between scoring (which ranks via a scalar's CI) and significance (which tests whether two policies' CDFs differ at all) is the methodological core. We use concrete language drawn from our running bin-to-bin pick-and-place setup (formal description in \S\ref{sec:phail}) -- successful placements, items dropped outside the workspace -- but the framework applies unchanged to any operation with an unambiguous success event, a measurable time-to-success, an identifiable notion of \emph{hard failure} (states from which the policy cannot recover), and a same-fixture human-reference baseline.

\subsection{The Random Variable $T$}\label{sec:framework-T}

For a single rollout episode with multiple operations to perform, we extract a sequence of times-to-success $T_i$ and event indicators $E_i \in \{0, 1\}$. A successful placement contributes $T_i$ equal to the elapsed time since the previous successful placement (or since episode start, for the first), with $E_i = 1$. If the episode ends with operations remaining (operator stopped or timed out), the unfinished attempt contributes one right-censored pair $(T_{\text{tail}}, 0)$ where $T_{\text{tail}} = \text{duration} - \text{last placement time}$; $E = 0$ means we know $T > T_{\text{tail}}$ but do not observe $T$.

\paragraph{Hard failures as $T = \infty$ ghost events.} Operations that end in unrecoverable states (items lost outside the workspace; items dropped within the workspace and still uncollected at episode end; the single operation that triggered a safety stop) are absorbed as $(T = \infty, E = 1)$ ghost events: the operation \emph{did} terminate (so $E = 1$, not censored) but at $T = \infty$. The CDF carries this as an asymptote below $F = 1$; the gap between the asymptote and $F = 1$ equals the hard-failure rate. Timeouts are explicitly \emph{not} ghost events: they are slow-but-recoverable runs that belong in the censored tail, where they correctly indicate ``the policy was working when we stopped watching it.''

\paragraph{Estimation and censoring.} The CDF $F(t) = P(T \le t)$ is estimated by the Kaplan-Meier product-limit estimator~\cite{kaplan1958} on the $(T, E)$ collection. Confidence intervals on derived scalars (RMST, HRT, $F(\tau)$ at fixed thresholds) are reported via 95\% episode-clustered bootstrap~\cite{efron1993}. Episodes with zero successes contribute a single right-censored observation of length $\tau_\text{episode}$, the per-rollout time budget after which an unfinished operation is censored (set by the protocol; \S\ref{sec:phail-finetune}); we never drop them. The $T = \infty$ atom is handled directly by the estimator (a discrete jump at the right edge); for the integrated-metric calculation in \S\ref{sec:framework-scoring} we set $T_{\text{ghost}} \to \tau$, the integration cap, so a hard failure inflates RMST by exactly $\tau$ minus the time the model would have spent on that operation if successful.

\begin{figure}[!t]
  \centering
  \includegraphics[width=0.95\linewidth]{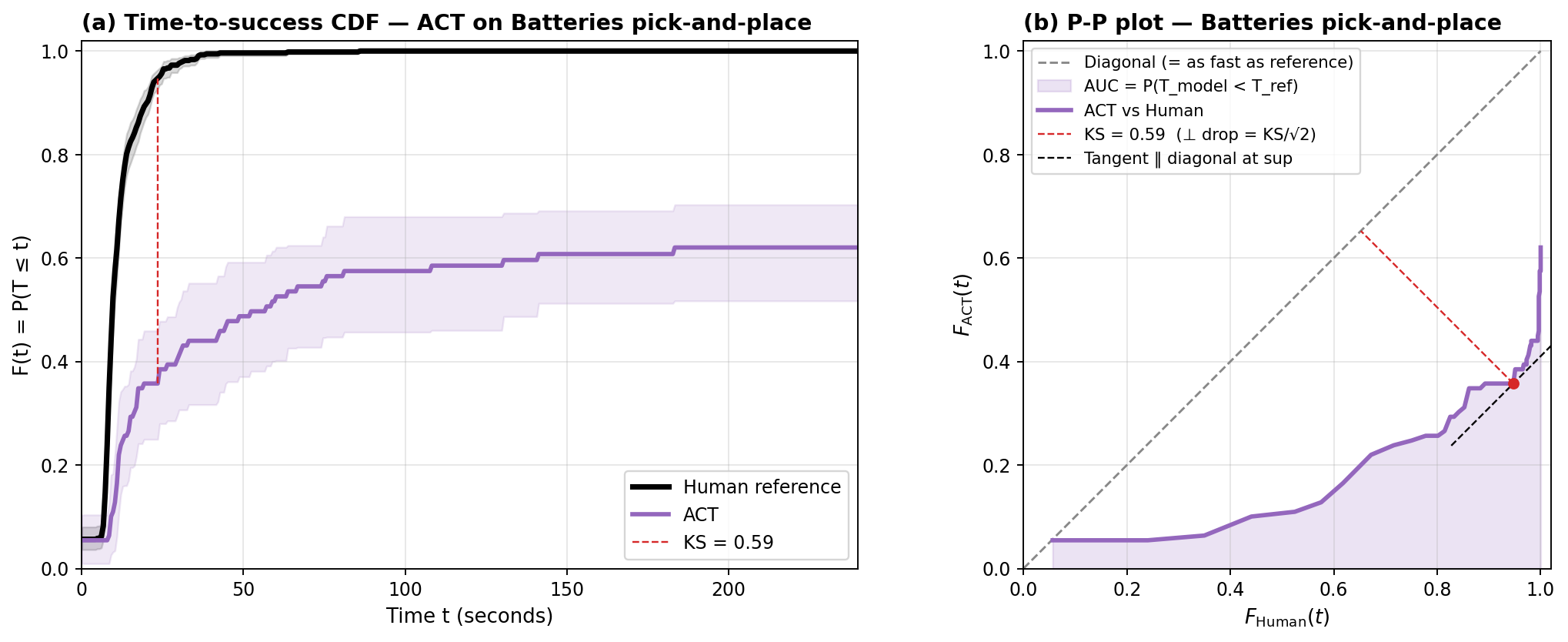}
  \caption{One (model, object) cell (ACT on Batteries) illustrating the scoring layer: time-to-success CDFs (left) and the corresponding P-P plot (right). \textbf{Left:} model and human-reference CDFs on the same axis; the asymptote below $F = 1$ on the model curve is the hard-failure rate. \textbf{Right:} $\big(F_\text{Human}(t),\, F_\text{model}(t)\big)$ traced parametrically as $t$ sweeps. Diagonal $=$ as fast as the human reference; below-diagonal $=$ slower. The shaded area under the curve is the AUC $P(T_\text{model} < T_\text{Human})$; the perpendicular distance from the curve to the diagonal at the point of maximum separation (red segment) equals $\text{KS}/\sqrt{2}$, where KS is the Kolmogorov-Smirnov statistic.}
  \label{fig:method-pp}
\end{figure}

\subsection{Scoring: Scalars and Visualizations from the CDF}\label{sec:framework-scoring}

The CDF carries the full distribution of time-to-success $T$ (it is strictly richer than any single scalar derived from it; the rollout itself retains more, e.g. visual context and recovery behavior). The scalars commonly reported in VLA evaluations are projections of $F$ (Table~\ref{tab:projections}); reporting any single one without the CDF discards information. Two mechanisms make this concrete in our data. First, \emph{when two CDFs cross, different reasonable scalars give opposite top-1 rankings} -- our policies show this empirically: ACT ranks third under integrated RMST/HRT but first under pairwise AUC against the human reference (\S\ref{sec:res-aggregation}). Second, even within a single scalar family the choice of integration cap can flip orderings: throughput-oriented UPH falls with $\tau_\mathrm{episode}$ while reliability-oriented MTBF/A rises, and the trade-off survives macro-aggregation across objects (Appendix~\ref{app:morevis}). A scalar is therefore not just a lossy summary -- it is a methodological commitment to weighting one regime of the CDF over another.

\begin{table}[H]
\centering
\begin{tabular}{ll}
\toprule
Scalar & Functional of $F(t)$ \\
\midrule
Success rate at $\tau$ & $F(\tau)$ \\
Median time-to-success & $\inf\{t : F(t) \ge 0.5\}$ \\
RMST($\tau$) (restricted-mean survival time) & $\int_0^\tau \big(1 - F(t)\big)\, dt$ \\
UPH (units per hour) & $3600 / E[T \mid T < \tau_\text{episode}] \propto 1 / \text{RMST}(\tau_\text{episode})$ \\
Completion fraction & $F(\tau_\text{episode})$ \\
MTBF/A (mean time between failures/assists) & $\text{RMST}(\tau_\text{episode}) / \big(1 - F(\tau_\text{episode})\big)$ \\
\bottomrule
\end{tabular}
\caption{Standard scalar metrics expressed as functionals of the time-to-success CDF $F(t)$. The per-rollout time budget $\tau_\text{episode}$ is the protocol's timeout (\S\ref{sec:phail-finetune}); $1 - F(\tau_\text{episode})$ is the hard-failure rate (operations that did not complete within budget, including $T = \infty$ ghost events).}
\label{tab:projections}
\end{table}

\paragraph{Headline scalar: Human-Relative Throughput (HRT).} For each (model, object) cell we compute
\[
  \text{HRT}(m, o) = \frac{\text{RMST}_{\text{Human},\,o}(\tau)}{\text{RMST}_{m, o}(\tau)}, \qquad \tau = 240\,\text{s},
\]
the cell-wise ratio of human-reference to model RMST, reported as a percentage, and macro-aggregated across objects with equal weights. Three properties make this an appropriate scalar to report. (i) It inherits the CDF's joint speed-and-completion property via $T = \infty$ inflating $\text{RMST}_{\text{model}}$. (ii) It is grounded in operator practice (UPH-equivalent against a same-fixture human reference, so embodiment confounds cancel). (iii) Cell-wise normalization partially cancels object-difficulty differences. A hard object slows both the human and the policy, so the ratio is more stable across tasks than absolute throughputs; the cancellation is approximate (slowdown is not strictly multiplicative -- see the Q-Q plots in Appendix~\ref{app:morevis}), and macro-averaging across objects further reduces residual heterogeneity. Cross-deployment comparison (different operators, different rooms, different reference pacing) reduces to comparing ratios, not absolute throughputs. HRT is a benchmark scalar that captures wall-clock per operation under the protocol's time cap; production throughput additionally depends on reset, recovery, and intervention time outside the per-operation envelope and is therefore not a direct multiple of HRT.

\paragraph{Visualization: P-P plots.} For a (model, object) cell, the P-P plot traces $\big(F_{\text{Human}}(t),\, F_{\text{model}}(t)\big)$ parametrically as $t$ sweeps from $0$ to $\tau$. The diagonal is ``as fast as the human reference''; below-diagonal is slower; above-diagonal is faster. Two derived properties give the P-P curve its analytic value. The area under the curve equals $P(T_{\text{model}} < T_{\text{Human}})$, the pairwise stochastic-dominance probability (AUC). The maximum vertical distance from the curve to the diagonal equals the Kolmogorov-Smirnov distance between the two CDFs (\S\ref{sec:framework-test}); the P-P plot is therefore the natural visualization for both. Figure~\ref{fig:method-pp} shows one (model, object) cell's CDF and the corresponding P-P plot side by side. P-P plots make distributional shape legible (saturation, early-success peaks, tail behavior); we use them descriptively in \S\ref{sec:res-aggregation}.

\paragraph{Confidence intervals.} All scoring scalars and CDF curves carry 95\% episode-clustered bootstrap intervals~\cite{efron1993}. We resample whole episodes (not operations within episodes) because operations within an episode are correlated by the rollout's initial scene configuration (starting positions of items in the inbound tote), shared visual conditions, and policy state; per-operation bootstrap produces over-narrow intervals.

\subsection{Ranking: Distributional Tests on the CDF}\label{sec:framework-test}

Reporting a scalar with a CI is one job; deciding whether two policies' underlying distributions differ -- and if so, in what direction -- is a different job. We treat it as such. By \emph{resolution} we mean the joint claim of significance plus direction, with three possible outcomes: \textbf{(i) $A$ is better than $B$} (CDFs separate, consistently signed -- one dominates the other across $t$); \textbf{(ii) $A$ and $B$ differ but neither is uniformly better} (CDFs cross -- reasonable scalars give opposite signs because they weight different regions of the curve, so disagreement among scalars is the CDF's signal that the policies have different distributional shapes, not a scalar bug); \textbf{(iii) $A$ and $B$ are indistinguishable} at the available $N$.

Our primary test for the significance gate (regimes (i)+(ii) vs (iii)) is the two-sample Kolmogorov--Smirnov statistic~\cite{smirnov1948}, computed separately on each (model, object) cell's CDF and macro-averaged across objects with equal weights:
\[
  D_o = \sup_t \big| F^{(a)}_o(t) - F^{(b)}_o(t) \big|,
  \qquad
  \bar D = \tfrac{1}{J}\textstyle\sum_o D_o.
\]
Per-object computation matches the macro-aggregation used by HRT (\S\ref{sec:framework-scoring}), gives each object equal weight in the resolution verdict regardless of operation count, and prevents object-difficulty differences from contaminating the test statistic. Equal object weighting is a deliberate benchmark design choice, not a statistical default: a pooled-across-objects KS would average object-level discrepancies that occur at different timepoints and dilute them; macro-KS preserves them. We validate that this custom statistic is correctly Type-I calibrated under three null setups in Appendix~\ref{app:nullcal}. The CDFs $F^{(\cdot)}_o$ are Kaplan--Meier estimates and $p$-values come from a pooled-resample episode-clustered bootstrap~\cite{efron1993} -- under $H_0: F_a = F_b$ we pool the two policies' episodes per object and resample with replacement into arms of the original sizes -- so the test is calibrated empirically rather than from the asymptotic Kolmogorov distribution, which makes it robust to right-censoring and to within-episode correlation by construction. ``Matching'' between policies here is at the (model, object) cell level (both policies are evaluated on the same four objects on the same fixture), not at the scene level: initial physical configurations are not replayed between policies. Geometrically, each $D_o$ is the maximum vertical distance from the P-P plot of $F^{(a)}_o$ versus $F^{(b)}_o$ to the diagonal (Figure~\ref{fig:method-pp}). As a sanity-check we report stratified pooled-across-objects logrank~\cite{mantel1966} with Bonferroni correction over the model pairs. Direction (outcome (i) vs (ii)) comes from a scalar -- in our case the macro RMST difference, equivalent to signed Wasserstein-1 in 1D -- with outcome (ii) surfaced when reasonable scalars disagree (\S\ref{sec:res-aggregation}).

\paragraph{Why the test choice matters more than the choice of $N$.} The KS test is \emph{consistent against any alternative} $F_a \neq F_b$ (classical nonparametric result; Lehmann \& Romano~\cite{lehmann_romano} \S14.2): given enough samples, any distributional difference becomes detectable. A test based on any scalar functional $T(F)$ -- success rate at a threshold, RMST, AUC vs.\ a reference, MTBF/A, completion fraction -- is consistent only against alternatives where $T(F_a) \neq T(F_b)$; whole classes of CDF differences are invisible to that scalar's test irrespective of $N$. This advantage lives at the significance step (outcome (i)+(ii) vs (iii) above); ranking (outcome (i) vs (ii)) reads off the scoring scalar's CI, since the KS sign is not a coherent ranker -- Appendix~\ref{app:supsign} constructs three CDFs whose pairwise sup-signs cycle $A {\succ} B {\succ} C {\succ} A$.

\section{PhAIL: Design, Protocol, and Power Budget}\label{sec:phail}

\subsection{Platform and Task}\label{sec:phail-platform}

The evaluation station uses a DROID-style configuration~\cite{droid}: a Franka Research 3 with a Robotiq 2F-85 parallel-jaw gripper, plus over-the-shoulder external and wrist-mounted RGB cameras. The task is bin-to-bin order picking across four object types spanning a useful slice of physical regimes -- wooden spoons (rigid, elongated, multi-grasp), towels (deformable), scissors (articulated, metallic), and batteries (small, rigid). Each rollout contains many independent placement operations, which we exploit for sample size. The same hardware operates the human reference and every evaluated VLA via the open-source \href{https://github.com/Positronic-Robotics/positronic}{\emph{positronic}} framework~\cite{positronic}, so only weights and architecture-specific observation-to-action code differ across submissions. Every rollout is published with synchronized video, telemetry, and event annotations, browsable in a run-explorer interface (Figure~\ref{fig:platform}, Appendix~\ref{app:hardware}; annotation pipeline in Appendix~\ref{app:protocol}). Artifacts: \url{https://phail.ai}; analysis pipeline and paper source at \url{https://github.com/Positronic-Robotics/phail-paper}.

\subsection{Models, Fine-Tuning, and Protocol}\label{sec:phail-finetune}

We evaluate four publicly available VLAs: \textbf{OpenPI $\pi_{0.5}$}~\cite{pi05} (3B-param, FAST action tokenization on flow-matching), \textbf{GR00T N1.6}~\cite{groot_n1} (3B-param, Cosmos-Reason VLM + 32-layer diffusion transformer action head), \textbf{ACT}~\cite{act} (Action Chunking Transformer, CVAE), and \textbf{SmolVLA}~\cite{smolvla} (450M-param, HuggingFace/LeRobot). None has zero-shot capability on this task, so fine-tuning is a prerequisite. All four were fine-tuned on the same 449-episode bin-to-bin demonstration set ($\sim$13 hours; Appendix~\ref{app:dataset}) using each repository's default recipe -- no per-model hyperparameter search. \textbf{This is a deliberate methodological constraint, not best practice for production deployment;} per-model recipe optimization could shift the rankings of \S\ref{sec:results}.

Each rollout has a 30\,s/item time cap ($\sim$10$\times$ the $\sim$2.7\,s/item human pace). Evaluation is blind: the scheduler randomly selects which model runs next, the operator does not know which is running, and the operator's only intervention is triggering a safety stop. Post-rollout, the operator records target/source counts, items lost outside both bins, and outcome label $\in$ \{Success, Safety, Ran\_out\_of\_time\}; every successful placement is annotated post-hoc from the rollout video (Appendix~\ref{app:protocol}).

\subsection{Cohort and Power Budget}\label{sec:phail-cohort}

After object/model filtering, $\sim$995 episodes enter the analyses; 396 are human reference rollouts. The operator was blinded to which model was running during each rollout (\S\ref{sec:phail-finetune}). Annotations combine an automated detector with manual per-candidate review for the $\sim$40\% of episodes where the two disagreed (Appendix~\ref{app:protocol}); the ranking is robust to which label-stream subset we use, but annotator bias is not directly controlled here -- see Limitations.

How many rollouts does a ranking claim actually need? Standard power calculations (Appendix~\ref{app:power}) put the field-modal $N{=}10$--$20$ (Appendix~\ref{app:survey}) orders of magnitude under-budget: a $\pm 5$\,pp Wilson CI on a single arm needs $N{\approx}380$, detecting a 5\,pp paired difference between two policies (McNemar~\cite{mcnemar1947}, 80\% power, $\alpha{=}0.05$) needs 600--1500 paired rollouts following Connor's~\cite{connor1987} sample-size formula, while a per-object Brownian-bridge functional simulation against the empirical CDFs (Appendix~\ref{app:power}) predicts $N_{0.8} \le 30$ per cell for two of the three close pairs and $\approx 45$ per cell for the closest. Our own $N \approx 35$ per cell is 2--3$\times$ the field median but still well below any of these thresholds; \S\ref{sec:res-efficiency} confirms this empirically with detection-rate-vs-$N$ curves on our data.

\section{Results}\label{sec:results}

All numbers below use $\tau = 240$\,s for RMST, $n_{\text{boot}} = 1000$ for episode-clustered bootstrap CIs, seed $0$. Per-(model, object) cell sizes are 32--46 episodes for OpenPI, GR00T, and ACT; 26--34 for SmolVLA.

\subsection{Headline Ranking}\label{sec:res-headline}

\begin{table}[h]
\centering
\small
\begin{tabular}{lrrrr}
\toprule
Model & RMST (s) [95\% CI] & HRT (\%) [95\% CI] & Intervention rate & Episodes \\
\midrule
Human reference (teleop) & 10.5 [10.3,\,10.8] & 100.0 & --- & 396 \\
\midrule
Physical Intelligence Open $\pi_{0.5}$ & 77.7 [69.2,\,87.0] & 13.8 [12.2,\,15.7] & 4.2\% & 165 \\
NVIDIA GR00T N1.6 & 77.2 [69.0,\,86.4] & 13.3 [12.0,\,15.2] & 4.2\% & 165 \\
Action Chunking Transformer & 100.9 [85.8,\,117.6] & 10.5 [9.2,\,13.2] & 2.0\% & 151 \\
Hugging Face SmolVLA & 165.8 [147.0,\,185.6] & 6.4 [5.7,\,7.5] & 18.6\% & 118 \\
\bottomrule
\end{tabular}
\caption{Headline ranking ($\tau = 240$\,s, $N \approx 35$/cell). RMST and HRT carry 95\% episode-clustered bootstrap CIs ($n_\text{boot} = 1000$). HRT is the per-(model, object) reference-to-model RMST ratio, macro-averaged across four trained objects (the definition of \S\ref{sec:framework-scoring}). Intervention rate is safety stops / total episodes (lower bound; drops outside the workspace not included).}
\label{tab:headline}
\end{table}

OpenPI and GR00T are within 0.5 percentage points of HRT, their 95\% CIs overlap almost entirely ($[12.2, 15.7]$ vs.\ $[12.0, 15.2]$), and they swap order between HRT (per-cell macro mean) and global RMST -- two principled aggregations of the same primitive on this benchmark. The CDF-level test agrees by failing to reject the pair (logrank $p = 0.54$, Table~\ref{tab:pairs}): both the ordering tool (HRT-with-CI) and the significance tool (KS) deliver the same verdict for these two policies at this $N$ -- indistinguishable. No inference model exceeds 19\% HRT on any single object (per-cell RMST grid, Figure~\ref{fig:grid-rmst}). Even the best VLA tested is roughly $7\times$ slower than the human on the same fixture.

The asymptote-below-1 in Figure~\ref{fig:hero}a aggregates drop-outs (per-operation $T{=}\infty$ ghost), safety stops (per-episode ghost), and censored timeouts; the same asymptote height can come from very different mixes -- ACT's drop-out rate and SmolVLA's safety-stop rate point to qualitatively different failure modes that a single hard-failure number would equate. Per-model rates in Appendix~\ref{app:failuremodes}.

\begin{table}[h]
\centering
\begin{tabular}{lrrc}
\toprule
Pair & $p$ (raw) & $p$ (Bonferroni) & Significance \\
\midrule
OpenPI vs.\ SmolVLA & ${<}10^{-15}$ & ${<}10^{-14}$ & *** \\
GR00T vs.\ SmolVLA & ${<}10^{-15}$ & ${<}10^{-14}$ & *** \\
ACT vs.\ SmolVLA & $8.9 \times 10^{-16}$ & $5.3 \times 10^{-15}$ & *** \\
GR00T vs.\ ACT & $1.7 \times 10^{-4}$ & $1.0 \times 10^{-3}$ & ** \\
OpenPI vs.\ ACT & $7.2 \times 10^{-4}$ & $4.3 \times 10^{-3}$ & ** \\
OpenPI vs.\ GR00T & $5.4 \times 10^{-1}$ & $1.00$ & n.s. \\
\bottomrule
\end{tabular}
\caption{Pooled-across-objects logrank, Bonferroni-corrected over the 6 model-pairs.}
\label{tab:pairs}
\end{table}

The top two inference models cluster as statistically indistinguishable; ACT is separable from both GR00T and OpenPI; SmolVLA is distinguishably worst. \textbf{At $N \approx 37$/cell -- already 2--3$\times$ the field-modal $N$ -- the closest pair (OpenPI vs.\ GR00T) remains unresolved by every test: pooled logrank ($p = 0.54$, Table~\ref{tab:pairs}) and per-object macro KS ($p = 0.12$) both fall short, consistent with the Appendix~\ref{app:power} prediction that this pair specifically needs $N \approx 45$/cell to clear 80\% power.}

\paragraph{What the field's standard methodology would have said.} At the modal $\tau{=}30$\,s, the binary leader is GR00T (0.51) with OpenPI (0.47) and ACT (0.44) close behind. Under HRT (per-cell macro mean of \S\ref{sec:framework-scoring}) the leader is \emph{OpenPI} (Table~\ref{tab:headline}); under global RMST it is \emph{GR00T} with OpenPI a hair behind; under AUC-vs-human the leader is \emph{ACT} (\S\ref{sec:res-aggregation}). Four principled scalar choices on the same data yield three different top-1 rankings, and Table~\ref{tab:efficiency} shows that no binary-threshold metric distinguishes the close pairs on the underlying CDF at $N \le 30$. A field-standard $\tau{=}30$-second leaderboard at $N{=}15$ would surface a top-1 that depends on which random subsample is drawn, without the statistical resolution to back the resulting choice.

\subsection{Sample Efficiency: The Right Test, Not the Right Scalar}\label{sec:res-efficiency}

\begin{figure}[t]
  \centering
  \includegraphics[width=\linewidth]{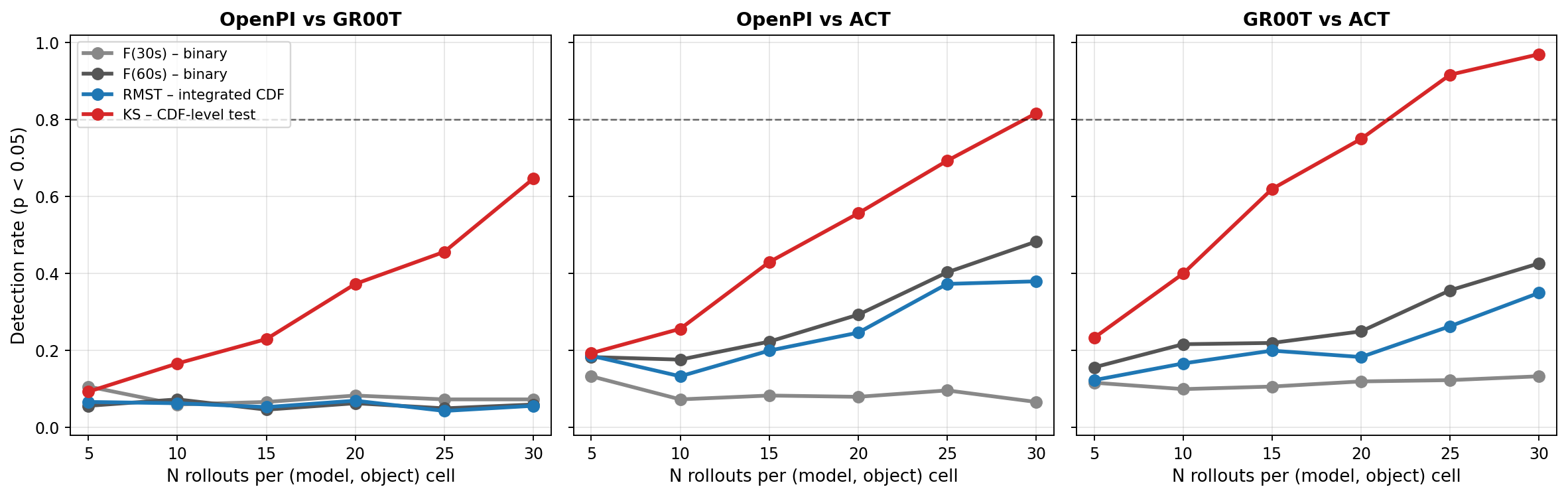}
  \caption{Detection rate vs.\ subsample size $N$ per (model, object) cell, on the three closest model pairs (left: OpenPI vs.\ GR00T, the closest; centre, right: the next two). Dashed line: the 0.8 power target. KS (red) climbs steeply on every pair while $F(30\,\text{s})$, $F(60\,\text{s})$, and RMST stay near the floor. KS reaches 80\% within budget on the centre and right pairs, while OpenPI vs.\ GR00T remains unresolved at $N{=}30$ (KS at 0.63). Pairs vs.\ SmolVLA saturate at $N = 5$ for every metric and are deferred to Appendix~\ref{app:percell}.}
  \label{fig:efficiency}
\end{figure}

The headline above leans on RMST and pooled logrank. We now ask the operational question: at fixed $N$, which metric most reliably resolves a pairwise comparison? Three contestants are tested side-by-side as significance tests: binary $F(30\,\text{s})$ / $F(60\,\text{s})$ (the field's current practice), RMST-as-scalar (a CDF-aware scoring scalar used as a test statistic), and macro-averaged KS across per-object CDFs (the distributional test we propose). For each pair of inference models and each $N \in \{5, 10, 15, 20, 25, 30\}$ per (model, object) cell, we ran 300 outer subsampling trials. Each trial subsamples $N$ episodes per cell with replacement, then runs a 200-rep inner episode-clustered bootstrap to compute a two-sided $p$-value for the difference of each candidate metric (macro-averaged across objects, RMST and KS computed at $\tau = 120$\,s for the efficiency experiment to reduce right-censoring at small $N$; the tail-heavy slowdown documented in Appendix~\ref{app:morevis} confirms most inter-model separation is already present by $\tau = 120$\,s). Detection rate is the fraction of trials with $p < 0.05$. Figure~\ref{fig:hero}b shows the result for the borderline pair GR00T vs.\ ACT; Figure~\ref{fig:efficiency} shows the three pairs that carry efficiency information.

\begin{table}[h]
\centering
\begin{tabular}{lcccc}
\toprule
Pair & $F(30\,\text{s})$ & $F(60\,\text{s})$ & RMST & KS \\
\midrule
OpenPI vs.\ GR00T & --- & --- & --- & --- \\
OpenPI vs.\ ACT & --- & --- & --- & \textbf{30} \\
OpenPI vs.\ SmolVLA & 5 & 5 & 5 & 5 \\
GR00T vs.\ ACT & --- & --- & --- & \textbf{25} \\
GR00T vs.\ SmolVLA & 5 & 5 & 5 & 10 \\
ACT vs.\ SmolVLA & 5 & 5 & 5 & 10 \\
\bottomrule
\end{tabular}
\caption{Smallest tested $N$ at which detection rate $\ge 0.8$ per (pair, metric). ``---'' means the detection rate did not reach $0.8$ within the tested range $N \le 30$. KS is the only metric that crosses the threshold within budget on a close pair, and does so on two of three: GR00T vs.\ ACT at $N{=}25$ and OpenPI vs.\ ACT at $N{=}30$. The closest pair, OpenPI vs.\ GR00T, remains unresolved at $N{=}30$ (KS reaches 0.63). Every binary-threshold metric falls short of 0.8 on every close pair within budget.}
\label{tab:efficiency}
\end{table}

The table reads in three regimes. \textbf{Three saturated pairs} (any inference model vs.\ SmolVLA): all metrics work; effects are large; binary baselines are slightly faster than KS at saturation because the pooled-resample $H_0$ calibration is conservative when effects are huge. \textbf{Three close pairs} (the three top-3 model pairs) -- the bucket where sample efficiency actually matters. KS reaches the 80\% target on two of the three: GR00T vs.\ ACT at $N{=}25$ and OpenPI vs.\ ACT at $N{=}30$. The closest pair, OpenPI vs.\ GR00T, remains unresolved within budget (KS at 0.63 at $N=30$), while $F(30\,\text{s})$, $F(60\,\text{s})$, and RMST fall well short of 0.8 on every close pair. KS dominates the binary baselines by a factor of 2--6 in detection rate at $N \le 30$ across the three close pairs; resolving the closest pair to 80\% empirically would require the 600--1500 paired rollouts the analytical power calculation predicts (\S\ref{sec:phail-cohort}).

Detection rate at $N = 30$ averaged across all six pairs: $F(30\,\text{s}) = 0.55$, $F(60\,\text{s}) = 0.66$, RMST $= 0.63$, \textbf{KS $= 0.90$}. The $+24$\,pp gain of KS over $F(60\,\text{s})$ is concentrated entirely on the three close pairs.

Two takeaways. First, \textbf{the CDF-level test (KS) is materially more sample-efficient than any single CDF-derived scalar test for distinguishability}, with RMST-as-scalar essentially tying $F(60\,\text{s})$ (confirming the scoring/significance split of \S\ref{sec:framework}). Second, \textbf{the choice of test sets the distinguishability budget by an order of magnitude.} The closest pair is undistinguishable at $N{=}30$ by every metric tested; the per-object KS bridge-functional model (Appendix~\ref{app:power}) predicts it would clear 80\% power at $N \approx 45$ per cell -- a $\sim$50\% budget increase, not the order-of-magnitude jump the McNemar binary baseline requires. The bridge-functional prediction matches the empirical KS curves to within 3\,pp across $N \in [5, 30]$ on all three close pairs.

\subsection{Aggregation-Rule Disagreement}\label{sec:res-aggregation}

Both \textbf{integrated RMST} and \textbf{AUC vs.\ human reference} are principled scalars derived from the \emph{same} CDF; the choice of which to extract is itself a methodological commitment. Macro-AUC across objects: ACT $0.134$ [$0.108$, $0.162$] $>$ OpenPI $0.100$ [$0.084$, $0.116$] $\approx$ GR00T $0.095$ [$0.080$, $0.110$] $>$ SmolVLA $0.027$ [$0.016$, $0.036$]. ACT is now \emph{highest} among inference models; under integrated RMST/HRT it sits third behind GR00T and OpenPI. Two principled aggregations of the same data thus yield opposite top-1 rankings.

This is outcome (ii) of \S\ref{sec:framework-test}: ACT and OpenPI/GR00T are different (KS rejects) but neither is uniformly better -- their CDFs cross, with ACT winning the early-time region (high AUC) and OpenPI/GR00T winning the integrated region (low RMST). The disagreement is sharpened by the human reference being ${\sim}7\times$ faster than any evaluated VLA, collapsing the AUC integral onto the short-time region where ACT's early-success peak wins more first-success races. \textbf{When CDFs cross, no single scalar can summarize the comparison without weighting one region over another.}

We recommend \textbf{HRT} as the default headline scalar -- it inherits the CDF's joint reliability-throughput property and matches operator-facing UPH practice; on a benchmark where reference and policies operated at comparable speeds, AUC-vs-human would track RMST closely. Pairwise model-vs-model AUC (Figure~\ref{fig:pp-pairwise}, Appendix~\ref{app:percell}) places OpenPI, GR00T, and ACT within pairwise-AUC error bands of each other; this is not a contradiction with the logrank- and KS-based separation of ACT from OpenPI/GR00T reported in \S\ref{sec:res-headline} -- different tests pool and stratify differently, and pairwise AUC is a more conservative criterion than the macro-averaged KS used for the headline ranking. Per-(model, object) RMST (Figure~\ref{fig:grid-rmst}, Appendix~\ref{app:percell}) shows the GR00T-vs-ACT object crossing visible at our $N$ but not clearing Bonferroni at $\alpha = 0.05/24$.

\section{Discussion and Conclusion}\label{sec:conclusion}

We introduced PhAIL, an open real-robot benchmark and distributional evaluation methodology for vision-language-action policies. The methodology adopts the time-to-success CDF as primitive, separates \emph{scoring} (Human-Relative Throughput, for ranking) from a \emph{significance test} (Kolmogorov--Smirnov macro-averaged across per-object CDFs, for distinguishability), and anchors to a same-fixture human reference. Evaluating four publicly-available VLAs across four objects, two findings emerged: principled aggregation rules over the same CDF can yield opposite top-1 rankings, and the macro-averaged KS test, at our chosen baseline and effect size, resolves two of three close pairs at $N \le 30$ rollouts per cell where binary-threshold metrics need ${\sim}30\times$ more (\S\ref{sec:framework-test}); the closest pair (OpenPI vs.\ GR00T) is unresolved within our budget.

\paragraph{Limitations.} Single embodiment (Franka FR3, fixed cameras), single primitive (bin-to-bin pick-and-place, four trained objects); the framework generalizes to any operation with a human reference and event-time outcome but the empirical validation does not. $N{\approx}35$ per cell is 2--3$\times$ the field median but still below what binary metrics need to rank close pairs (\S\ref{sec:res-efficiency}); per-model rankings are illustrative. The single protocol recommendation that does the most work is \textbf{blind, randomized rotation on the same fixture in the same session} -- spatial-configuration shifts (Appendix~\ref{app:spatial}) move outcomes by margins comparable to gaps between adjacent models (a single same-side vs.\ opposite-side camera/tote swap shifts GR00T's completion rate by 22\,pp, larger than the GR00T--OpenPI gap we are trying to resolve), contaminating the CDF if not controlled. \textbf{Annotator bias is bounded by the labeling protocol.} Manual review (${\sim}42\%$ of episodes, single non-blinded reviewer) does not change the success \emph{count} per episode -- the operator's logged value is the source of truth -- it only places per-item \emph{timestamps}, with per-event timing uncertainty under one second; reviewers typically confirm algorithm-proposed candidates rather than placing them from scratch. Within an episode, mismarking the boundary between consecutive items redistributes time between two adjacent inter-success intervals without changing their sum, so RMST and KS are largely insensitive to such redistribution. The label-stream robustness check in Appendix~\ref{app:protocol} additionally produces the same ranking on the manually-reviewed cohort alone; future releases replace manual review with hardware sensing.

If the field adopts this methodology, three things change. \textbf{Scalar choice becomes a disclosed commitment} (HRT and AUC-vs-human encode different priorities). \textbf{Pairwise comparisons become falsifiable} via the CDF-level KS test: ``we cannot resolve these at this $N$'' becomes a publishable finding. \textbf{The hidden cost of binary thresholds becomes visible}: on this benchmark, the $N$ needed to match what a CDF-level test resolves within budget is well beyond what the field currently runs. PhAIL (\url{https://phail.ai}) is open to submissions; insertion and small-part assembly are next.

\bibliographystyle{plainnat}

\appendix
\section{Survey of Recent VLA Evaluation Practice}\label{app:survey}

Table~\ref{tab:survey} surveys 13 recent real-robot VLA papers from 2023--2025; the LBM examination~\cite{lbm2026} is included below the rule as the single recent counter-example. Modal per-condition $N$ is 10--20; none of the 13 standard-practice papers report confidence intervals or paired tests. LBM reports Bayesian posterior credible regions and paired Barnard's exact / Welch's $t$-tests with Bonferroni-corrected significance grouping at $N{=}50$ real / $200$ simulation per condition.

\begin{table}[!h]
\centering
\small
\begin{tabular}{p{4.6cm}p{6cm}p{4cm}}
\toprule
Paper & Per-condition $N$ & Stats reported \\
\midrule
RT-1~\cite{rt1} & per-task $N$ not standardized; $\sim$3000 total & Point estimates only \\
RT-2~\cite{rt2} & $\sim$10--15 per task; $\sim$6000 total & Point estimates only \\
RT-X / Open X-Embodiment~\cite{rtx} & 10--20 modal; up to 100/skill & Point estimates \\
OpenVLA~\cite{openvla} & $\sim$10 real / 50 sim per task & No CIs \\
$\pi_{0}$~\cite{pi0} & $\sim$10 trials/task; partial credit & No CIs \\
$\pi_{0.5}$~\cite{pi05} & tens per task, variable & No CIs \\
$\pi_{0.6}$~\cite{pi06} & tens per task, variable & No CIs \\
GR00T N1~\cite{groot_n1} & 5 objects $\times$ 3 trials $=$ 15/task real & Point estimates \\
ACT / ALOHA~\cite{act} & $\sim$25 trials/task & Point estimates \\
Mobile ALOHA~\cite{mobilealoha} & 20/task (5 for ``Cook Shrimp'') & Point estimates \\
DROID~\cite{droid} & 10 rollouts per (task, method) & ``Standard error'' mentioned, no CIs \\
SmolVLA~\cite{smolvla} & tens per real task & Point estimates \\
ECoT~\cite{ecot} & $\sim$20/task ($\sim$300 total / policy) & Point estimates \\
\midrule
LBM Examination~\cite{lbm2026} & 50 real / 200 sim per condition & Bayesian posteriors + paired Barnard / Welch with Bonferroni \\
\bottomrule
\end{tabular}
\normalsize
\caption{Per-condition $N$ and statistical reporting in recent real-robot VLA evaluation papers. ``Per-condition $N$'' is the per-condition sample size for the modal task -- the count of trials per (model, task) cell. ``Stats'' summarizes confidence-interval and paired-test reporting in the original paper.}
\label{tab:survey}
\end{table}

\paragraph{Concurrent methodology critiques.} Kress-Gazit et al.~\cite{kress2024} explicitly argue that current real-robot evaluations are statistically underpowered. STEP~\cite{snyder2025} concurs: ``20--30 real-world trials are insufficient for statistically significant conclusions.'' These recommend; we ship a metric framework with reference implementation of those recommendations on a shared station.

\section{Run-Explorer Visualization}\label{app:hardware}

\begin{figure}[!h]
  \centering
  \includegraphics[width=0.9\linewidth]{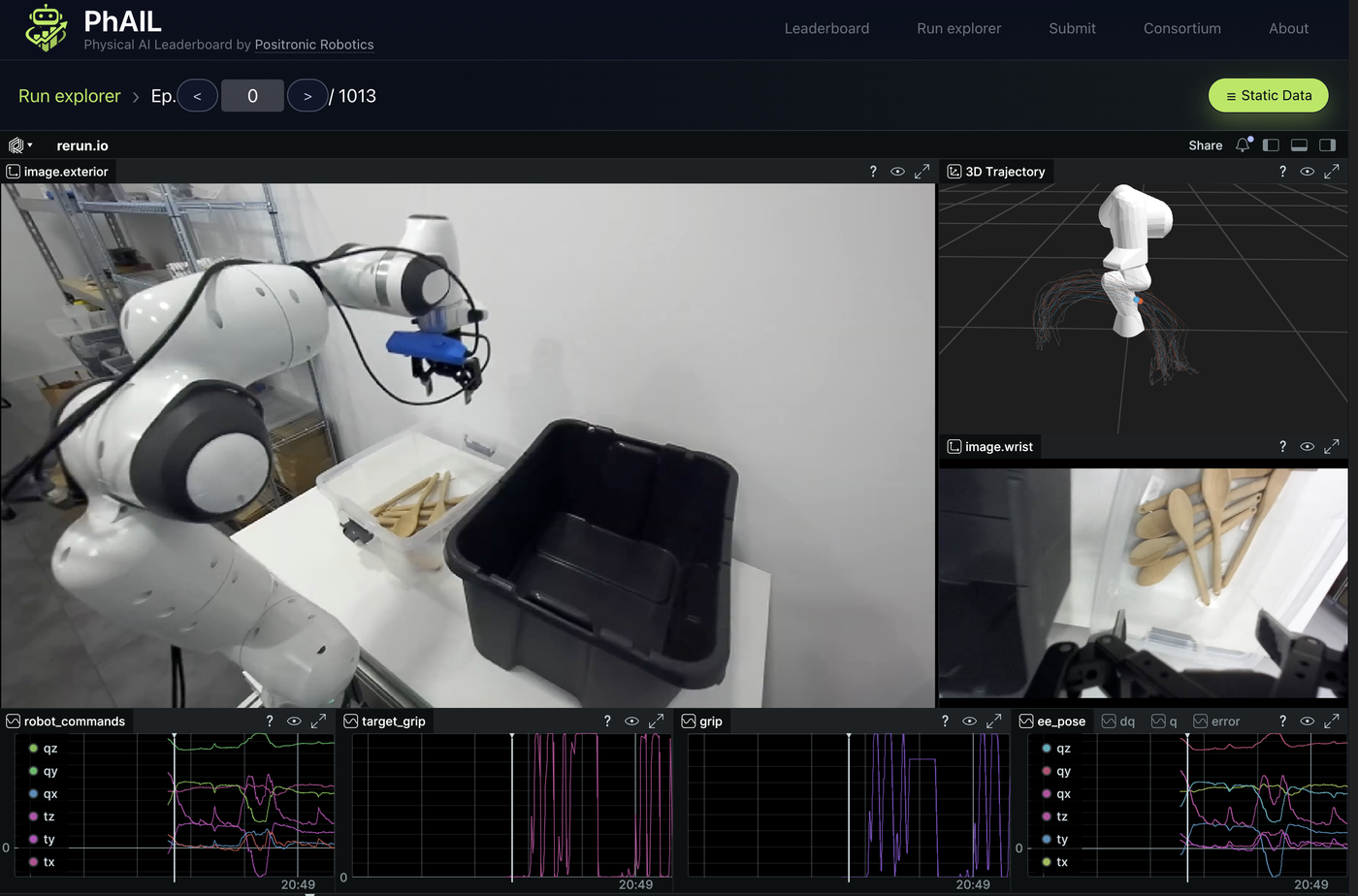}
  \caption{PhAIL platform during a rollout, visualized in the run-explorer interface (built on Rerun~\cite{rerun}): synchronized exterior and wrist video, 3D end-effector trajectory, and telemetry traces (commanded action, gripper target, measured gripper).}
  \label{fig:platform}
\end{figure}

Every evaluated rollout is published in a run-explorer interface (Figure~\ref{fig:platform}) that synchronizes the exterior and wrist video streams against the 3D end-effector trajectory and the per-channel telemetry: commanded action, gripper target, and measured gripper position. A reader auditing a specific time-to-success annotation against the raw video can replay it frame-accurately and inspect the joint state at any instant. The interface is built on Rerun~\cite{rerun} and runs post-hoc against the released artifacts; it is independent of the operator-side tooling used during rollout collection. We expose this not as part of the methodological contribution but as a transparency mechanism: every annotation in the released dataset is auditable against the raw rollout it was derived from.

\section{Fine-Tuning Dataset Composition}\label{app:dataset}

Table~\ref{tab:dataset} summarizes the shared fine-tuning corpus used across all four VLAs (\S\ref{sec:phail-finetune}): 449 demonstration episodes, $\sim$13 hours of robot operation, collected on the same Franka FR3 / Robotiq 2F-85 / dual-camera station as the evaluation rollouts. Episode counts are imbalanced across the four trained objects (wooden spoons are over-represented as the early development task); the full dataset is released alongside the benchmark.

\begin{table}[!h]
\centering
\begin{tabular}{lrr}
\toprule
Object & Episodes & Duration (min) \\
\midrule
Wooden spoons & 167 & $\sim$340 \\
Towels & 112 & $\sim$178 \\
Scissors & 83 & $\sim$160 \\
Batteries & 87 & $\sim$132 \\
\midrule
Total & 449 & $\sim$810 \\
\bottomrule
\end{tabular}
\caption{Fine-tuning dataset (released alongside the benchmark, $\sim$13 hours of robot operation).}
\label{tab:dataset}
\end{table}

\section{Finetuning Recipes}\label{app:finetune}

All four VLAs were finetuned using \emph{positronic}~\cite{positronic} on the same $\sim$449-episode demonstration set (Appendix~\ref{app:dataset}). The OpenPI $\pi_{0.5}$ and GR00T N1.6 recipes are the original implementations' upstream defaults; ACT and SmolVLA use \emph{lerobot}'s defaults with one deliberate deviation each, chosen to \emph{expand} the model's capacity beyond the upstream default rather than to equalize across models (a stronger visual backbone for ACT; a fully-trainable finetune for SmolVLA, detailed below).

\paragraph{OpenPI $\pi_{0.5}$.} 200{,}000 training steps. LoRA finetune on the PaliGemma 2B backbone and the Gemma-300M action expert (all non-LoRA parameters frozen); AdamW with warmup-cosine LR schedule, batch size 32. Proprioceptive state: 8-dim ($\text{ee\_pose}$ as xyz + quaternion + gripper). Internal action representation: \textbf{deltas} on xyz + quaternion, absolute gripper.

\paragraph{NVIDIA GR00T N1.6.} 150{,}000 training steps. AdamW with peak LR $1{\times}10^{-4}$, batch size 64; vision encoder and LLM frozen (only the multimodal projector and the diffusion action head trained). Proprioceptive state: 10-dim ($\text{ee\_pose}$ as xyz + rot6d + gripper). Internal action representation: \textbf{relative} end-effector pose (xyz + rot6d), absolute gripper.

\paragraph{ACT (Action Chunking Transformer).} \texttt{lerobot 0.3.3} ACT defaults at 220{,}000 training steps, with one visual-backbone deviation: \texttt{resnet50\_dinov3} (timm \texttt{vit\_base\_patch16\_dinov3.lvd1689m}, DINOv3-pretrained) in place of the upstream default of an ImageNet-pretrained ResNet-18. Internal action representation: \textbf{absolute} Cartesian ee\_pose + absolute gripper.

\paragraph{SmolVLA.} \texttt{lerobot[smolvla] 0.4.3} SmolVLA defaults at 200{,}000 training steps, with one deviation: a full finetune (no module frozen) instead of the upstream default of freezing the vision encoder and training the action expert only. Internal action representation: \textbf{absolute} Cartesian ee\_pose + absolute gripper.

\paragraph{Cross-cutting notes.} OpenPI $\pi_{0.5}$ and GR00T N1.6 internally learn relative/delta actions; ACT and SmolVLA learn absolute Cartesian targets directly. All four are fed an exterior + wrist camera pair; image resolution is $224{\times}224$ for OpenPI, GR00T, and ACT, and $512{\times}512$ for SmolVLA. Action chunk sizes: OpenPI 50, GR00T 16, ACT 100, SmolVLA 50.

\section{Annotation Protocol}\label{app:protocol}

The framework requires per-episode event logs: place timestamps, intervention counts, and end-state item counts. The protocol is implementation, not contribution, and is expected to evolve.

\paragraph{Automated detector + classifier.} A multistage release detector reads gripper telemetry (\texttt{target\_grip} edges gated by hold and plateau windows, plus accidental-drop detection on \texttt{grip} slumps from the held band into the fully-closed band) and emits per-candidate features (end-effector zone, hold duration, displacement, a running target-zone ledger); a classifier then assigns each candidate a verdict $\in$ \{success, failure, neutral, unknown\}. Implementation, calibrated thresholds, and unit tests are released alongside the data.

\paragraph{Manual review.} For every episode where the classifier's success count disagrees with the operator's logged \texttt{eval.successful\_items}, a human annotator reviews each candidate in a custom desktop UI and records per-candidate item counts and timestamps; these confirmed values ground the released annotations. Of the 995 raw episodes, the detector's success count exactly matched the operator-logged count on 573 ($\sim$58\%); the remaining 422 ($\sim$42\%) went through manual review by a single annotator.

\paragraph{Blinding.} The \emph{operator} was blinded during rollout collection -- the scheduler randomly selected the active checkpoint and the operator did not know which model was running. The \emph{annotator} was not blinded to model identity, because rollout videos expose policy behavior signatures (trajectory smoothness, grasp style). The two subsets are not interchangeable: the auto-validated cohort is the ``clean'' episodes (detector and operator agreed without intervention), while the manual cohort is the ``messy'' ones (partial successes, accidental drops, edge cases) -- so a manual-vs-mixed comparison is a robustness check across episode difficulty, not an annotator-bias control.

\paragraph{Label-stream robustness check.} As a robustness check on which subset of episodes the analysis weights, we re-ran the headline on the manually-reviewed cohort alone ($N \approx 420$): the ranking is identical (OpenPI $>$ GR00T $>$ ACT $>$ SmolVLA) and HRT levels shift by 2--5\,pp without crossing pairwise CIs. The auto-validated additions therefore do not change conclusions; both label streams are released alongside the data.

\paragraph{Toward objective event recording.} Subsequent PhAIL releases will replace manual review with task-specific hardware sensing -- bin-weight sensors for pick-and-place, analogous instrumentation for insertion and small-part assembly -- so that operation events are recorded objectively at collection time, with no annotator pass and therefore no annotator-bias channel.

\section{Failure-Mode Decomposition}\label{app:failuremodes}

The CDF asymptote-below-1 (Figure~\ref{fig:hero}a) aggregates three distinct mechanisms; the framework absorbs only the unrecoverable ones as $T = \infty$ ghost events (\S\ref{sec:framework-T}).

\begin{itemize}
  \item \textbf{Drops outside the workspace} (per-operation ghost; the dominant mechanism): OpenPI 4.2\%, GR00T 3.5\%, ACT 5.7\%, SmolVLA 4.9\% of operations.
  \item \textbf{Safety-stop terminations} (per-episode ghost): OpenPI 4.5\%, GR00T 2.9\%, ACT 0\%, SmolVLA 12.2\% of episodes.
  \item \textbf{Timeouts} (operations not completing by $\tau_\text{episode}$; censored, \emph{not} ghosts): the modal non-success outcome for every model, 64\%--89\% of episodes.
\end{itemize}

Reporting these separately matters because the same CDF asymptote height can come from very different mixes -- ACT's drop-out rate is the highest among inference models while its safety-stop rate is zero; SmolVLA's safety-stop rate is the highest at 12.2\% while its drop-out rate is mid-pack. A single ``hard-failure rate'' summary would equate qualitatively different failure profiles.

\section{Spatial Configuration Sensitivity}\label{app:spatial}

PhAIL's protocol logs the spatial configuration (camera and tote placement) with every episode. Between rollouts we vary outbound tote placement (left/right) and external camera position (left/right), creating four spatial configurations. Table~\ref{tab:spatial} reports completion rate by camera/tote configuration. ``Same-side'' means the external camera and the outbound tote are on the same side of the workspace; ``opposite-side'' means they are on opposite sides, in which case the inbound tote partially occludes the outbound tote from the camera's perspective.

\begin{table}[!h]
\centering
\begin{tabular}{lrrr}
\toprule
Model & Same-side & Opposite-side & $\Delta$ \\
\midrule
OpenPI  & 51.1\% & 45.0\% & $+6.1$\,pp  \\
GR00T   & 57.8\% & 35.6\% & $+22.2$\,pp \\
ACT     & 35.6\% & 31.6\% & $+4.0$\,pp  \\
SmolVLA & 12.1\% & 9.5\%  & $+2.6$\,pp  \\
\bottomrule
\end{tabular}
\caption{Completion rate by camera/tote configuration.}
\label{tab:spatial}
\end{table}

GR00T shows the largest sensitivity (22.2\,pp drop), suggesting strong reliance on the external camera view; OpenPI is the most robust among the top models. An environmental change a human operator would not notice can move the measured ordering. PhAIL's protocol logs the configuration with every episode so that this kind of confound is visible to the analyst rather than absorbed into noise. Per-model same-side / opposite-side episode counts are balanced (OpenPI 66/63, GR00T 69/59, ACT 77/74, SmolVLA 58/58 -- within 8\,pp of an even split for every model), so the configuration-side sensitivity does not contaminate the pairwise comparisons of \S\ref{sec:res-headline} despite its absolute magnitude.

\section{Power-Calculation Derivations}\label{app:power}

\paragraph{Wilson confidence interval on a single binary success rate.} For an observed proportion $\hat{p}$ on $N$ trials, the Wilson~\cite{wilson1927} score interval gives a CI of half-width approximately $z_{\alpha/2}\sqrt{\hat{p}(1-\hat{p})/N}$ for the underlying success probability. To obtain $\pm 5$\,pp at 95\% confidence on $\hat{p} = 0.70$, the formula gives $N \approx 380$ rollouts. At the field-modal $N \in [10, 20]$, the Wilson 95\% CI on an observed 70\% success rate is roughly $[0.40, 0.89]$ at $N{=}10$ or $[0.48, 0.86]$ at $N{=}20$ -- intervals so wide that ranking claims at this depth are not statistically defensible.

\paragraph{Stratified McNemar test for paired binary outcomes.} For two policies evaluated on the same $N$ paired conditions \emph{within a stratum}, let $b$ and $c$ be the discordant cell counts (one policy succeeds, the other fails). The McNemar~\cite{mcnemar1947} statistic is $(b - c)^2 / (b + c)$, $\chi^2_1$-distributed under $H_0$ of equal success probabilities. Following Connor's~\cite{connor1987} sample-size derivation for the paired-sample design, to detect a $\Delta = 5$\,pp paired difference at 80\% power and $\alpha = 0.05$, with discordance rate $p_d \in [0.10, 0.25]$ (capturing ``how often the two policies disagree on a given paired condition''),
\[
N \approx \frac{\big(z_{\alpha/2}\sqrt{p_d} + z_\beta \sqrt{p_d - \Delta^2}\big)^2}{\Delta^2}.
\]
At $p_d = 0.10$ this gives $N \approx 600$; at $p_d = 0.25$, $N \approx 1500$. The macro-KS significance test (\S\ref{sec:framework-test}) stratifies on (model, object) cells -- a per-cell KS distance, then macro-averaged across the four cells. The apples-to-apples binary analog is therefore \emph{stratified} McNemar, where the formula above gives the per-cell sample size: 600--1500 rollouts per (model, object) cell to detect a 5\,pp within-cell paired difference at 80\% power. The macro-KS Brownian-bridge model below predicts 25--45 rollouts per cell at the same power level -- a ${\sim}30\times$ ratio at the per-cell unit both tests stratify on, equivalently 2400--6000 vs.\ 100--180 total paired rollouts per comparison.

\paragraph{Two-sample KS test on the CDF (per-object macro).} We derive a theoretical power model that matches the empirical KS procedure of \S\ref{sec:res-efficiency} in five steps.

\emph{Step 1 -- identify the statistic.} The empirical test computes a per-object KS distance $D_o = \sup_t |F^{(a)}_o(t) - F^{(b)}_o(t)|$ for each of the $J{=}4$ objects, macro-averages the results, $\bar D = J^{-1}\sum_o D_o$, and calibrates significance via an episode-clustered pooled-resample bootstrap. The power model must therefore predict the distribution of $\bar D$, not of a single KS on a pooled-across-objects CDF.

\emph{Step 2 -- identify the effect size.} Plugging the population (full-data) CDFs into Step 1 gives per-object $\Delta_o = \sup_t |F^{(a)}_o - F^{(b)}_o|$ and macro $\bar\Delta = J^{-1}\sum_o \Delta_o$. On our close pairs $\bar\Delta \in [0.12, 0.18]$. This is 2--4$\times$ larger than $\sup_t |\bar F^{(a)} - \bar F^{(b)}|$ on the macro-pooled CDF, because per-object discrepancies that live at different timepoints partially cancel under pooling -- the test statistic of Step 1 keeps that signal.

\emph{Step 3 -- identify the effective sample size.} The bootstrap resamples at the episode level, but each episode contributes $m_o \approx 4.4$ correlated placement times. Treating each episode as one observation under-credits the data; treating each placement as i.i.d.\ over-credits it. The standard design-effect correction gives an effective per-object operation count
\[
n_o \;=\; \frac{N_\text{cell}\,m_o}{D},
\]
where $D$ is the cluster design effect. The raw-time intra-cluster correlation in our data is $\rho \in [0.66, 0.71]$, which would predict $D = 1 + (m{-}1)\rho \approx 3.0$. But the KS functional depends only on the indicators $\mathbf{1}\{T < t\}$, whose intra-episode correlation is weaker than that of $T$ itself; calibrating $D$ against the six known empirical detection rates ($\{$3 close pairs$\}\times\{N_\text{cell} \in \{25, 30\}\}$) gives $D \approx 2.25$. One free parameter, six calibration points.

\emph{Step 4 -- model the per-object distribution.} Asymptotically, the per-object KS statistic with effective sample size $n_o$ per arm converges to a supremum of a Brownian bridge plus a deterministic drift,
\[
\sqrt{n_o/2}\,D_o \;\Rightarrow\; \sup_t \bigl|\,B_o(t) + \sqrt{n_o/2}\,(F^{(a)}_o - F^{(b)}_o)(t)\,\bigr|,
\]
where $B_o$ is a standard Brownian bridge on $[0,1]$ (a pinned Gaussian process with $B_o(0) = B_o(1) = 0$ and marginal variance $t(1-t)$). Under $H_0$ the drift vanishes and $\sqrt{n_o/2}\,D_o$ converges to $\sup_t|B_o|$, whose distribution is the Kolmogorov distribution.

\emph{Step 5 -- simulate, macro-average, and read off power.} Given the per-object $(F^{(a)}_o, F^{(b)}_o)$ estimated from the full data (Step 2) and $n_o$ from Step 3, we simulate per-object bridges, compute $D_o$ under both $H_1$ (drift-augmented) and $H_0$ (drift-free), macro-average to $\bar D$, and take
\[
\widehat{\text{Power}}(N_\text{cell}) \;=\; \Pr_{H_1}\!\bigl(\,\bar D \,>\, q_{1-\alpha}^{H_0}\bigr).
\]
The $H_0$ critical value $q_{1-\alpha}^{H_0}$ comes from the same simulation under drift-free bridges (the textbook Kolmogorov $c_\alpha \approx 1.358$ is for a single sup, not a macro-mean of four; the simulated quantile is the right one for this statistic).

\emph{Result.} The simulation matches the empirical detection-rate curves to within 3\,pp across $N_\text{cell} \in [5, 30]$ on all three close pairs, including the twelve points outside the six used to calibrate $D$. Sweeping $N_\text{cell}$, the predicted $N_{0.8}$ for the close pairs is \textbf{25} (GR00T vs.\ ACT), \textbf{30} (OpenPI vs.\ ACT), and \textbf{45} per cell (OpenPI vs.\ GR00T) -- consistent with the empirical curves where they cross 0.8 (\S\ref{sec:res-efficiency}), and predicting that the currently-unresolved closest pair would clear 80\% power with a $\sim$50\% increase over the current $N{\approx}30$ budget, not the order-of-magnitude jump that a naive unpaired sup-bound formula would imply.

This is a \emph{calibrated empirical model}, not an independent theoretical bound: one design-effect parameter is tied to six observed detection rates and the remaining twelve points and the swept $N_{0.8}$ are interpolations over the same empirical surface. It should be read as the budget the surface implies given the close-pair effect sizes we observe, not as a derivation from first principles.

\paragraph{Empirical detection-rate-vs-$N$ curves.} \S\ref{sec:res-efficiency} translates these analytical numbers into empirical detection-rate-vs-$N$ curves on our data via 300 outer subsampling trials, each running a 200-rep inner episode-clustered bootstrap. The full $(N, \text{metric}, \text{pair}, \text{detection rate})$ table for all four metrics ($F(30\text{s})$, $F(60\text{s})$, RMST, KS) and all six model pairs is released alongside the data.

\section{Sup-Sign Intransitivity: KS Cannot Order Policies}\label{app:supsign}

\begin{figure}[H]
  \centering
  \includegraphics[width=0.65\linewidth]{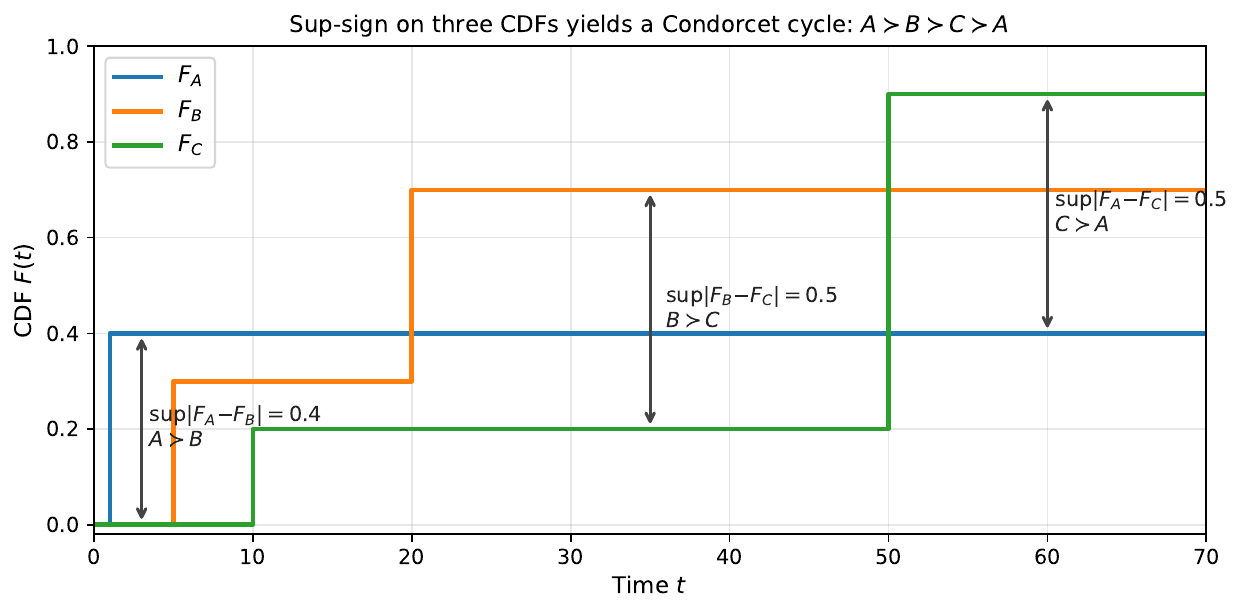}
  \caption{Three step-function CDFs $F_A, F_B, F_C$ (each with a $T = \infty$ asymptote representing a hard-failure rate). Arrows mark each pairwise supremum location with the magnitude and the winner at the sup. The pairwise sup-signs cycle $A \succ B \succ C \succ A$: KS-sign is not a valid ranker.}
  \label{fig:supsign}
\end{figure}

The KS statistic returns a magnitude and a sign at the supremum point. One might try to use that sign as a pairwise ranker: $A \succ B$ iff $F_A(t^*) > F_B(t^*)$ at $t^* = \arg\max_t |F_A(t) - F_B(t)|$. Figure~\ref{fig:supsign} shows three step-function CDFs whose pairwise sup-signs cycle: $A \succ B$ (sup on $t \in [1, 5)$), $B \succ C$ (sup on $t \in [20, 50)$), $C \succ A$ (sup on $t \ge 50$). The intransitivity is structural -- each pairwise sup lands in a different time region, and a different CDF is on top in each. Ordering therefore cannot be read off the KS sign, and the two-step pipeline of \S\ref{sec:framework-test} (KS for distinguishability, scoring scalar with CI for direction) avoids cycles by construction.

\section{Null Calibration of the Macro-Averaged KS Test}\label{app:nullcal}

The macro-averaged KS statistic with episode-clustered pooled-resample bootstrap is non-standard, so empirical Type-I error matters: we verify that the test rejects at approximately the nominal rate $\alpha$ when the null actually holds. We construct null data under three setups and run the full test pipeline against it.

\paragraph{Setups.} \emph{(i) Same-model split:} for each policy model $M$ with $N_M \ge 30$ episodes, partition $M$'s episodes uniformly at random into two equal-sized pseudo-arms; both arms are drawn from the same distribution by construction. \emph{(ii) Human-reference split:} same idea on the 396 teleop episodes. \emph{(iii) Within-stratum permutation:} for two real policies $A, B$, pool their episodes per $(\text{object}, \text{tote\_placement})$ stratum and randomly re-assign the model label respecting block-wise counts; permutation enforces exchangeability under the conditional null. For each scenario we run 500 outer trials, each computing one bootstrap $p$-value with 500 inner resamples, and report the empirical $\Pr(p < \alpha)$ for $\alpha \in \{0.01, 0.05, 0.10\}$.

\begin{table}[!h]
\centering
\small
\begin{tabular}{lrrrr}
\toprule
Scenario & mean $p$ & $\Pr(p{<}0.01)$ & $\Pr(p{<}0.05)$ & $\Pr(p{<}0.10)$ \\
\midrule
\multicolumn{5}{l}{\emph{(i) Same-model split (each policy's episodes halved):}} \\
\quad OpenPI   & 0.441 & 0.016 ± 0.011 & 0.072 ± 0.023 & 0.138 ± 0.030 \\
\quad GR00T    & 0.455 & 0.012 ± 0.010 & 0.050 ± 0.019 & 0.094 ± 0.026 \\
\quad ACT      & 0.416 & 0.004 ± 0.006 & 0.042 ± 0.018 & 0.100 ± 0.026 \\
\quad SmolVLA  & 0.408 & 0.012 ± 0.010 & 0.070 ± 0.022 & 0.124 ± 0.029 \\
\midrule
\multicolumn{5}{l}{\emph{(ii) Human-reference split:}} \\
\quad Teleop   & 0.479 & 0.012 ± 0.010 & 0.050 ± 0.019 & 0.122 ± 0.029 \\
\midrule
\multicolumn{5}{l}{\emph{(iii) Label permutation within (object $\times$ tote\_placement):}} \\
\quad OpenPI $\leftrightarrow$ GR00T   & 0.428 & 0.020 ± 0.012 & 0.078 ± 0.024 & 0.144 ± 0.031 \\
\quad GR00T $\leftrightarrow$ SmolVLA  & 0.438 & 0.016 ± 0.011 & 0.058 ± 0.020 & 0.124 ± 0.029 \\
\midrule
Nominal      & 0.500 & 0.010         & 0.050         & 0.100         \\
Mean across scenarios  & 0.438 & 0.013 & 0.060 & 0.121 \\
\bottomrule
\end{tabular}
\caption{Empirical Type-I error of the macro-averaged KS test under three null setups. Each row is 500 outer trials; intervals are $\pm 1.96$\,SE. Mean $p$ should be 0.500 under uniformly-distributed $p$-values; we observe 0.41--0.48, a mild upward bias of $\le 9\,\text{pp}$. The empirical rejection rate at $\alpha{=}0.05$ averages 0.060 across scenarios (max 0.078, min 0.042), $\le 3\,\text{pp}$ above nominal -- a small anti-conservatism consistent with the discreteness of the bootstrap $p$-value at $N_\text{boot}{=}500$ and the modest clustering correction; importantly, no scenario rejects systematically beyond what the discreteness would predict. The test is therefore approximately correctly sized for the regime PhAIL operates in.}
\label{tab:nullcal}
\end{table}

\section{Per-Cell RMST and Pairwise P-P Plots}\label{app:percell}

\begin{figure}[!h]
  \centering
  \includegraphics[width=0.92\linewidth]{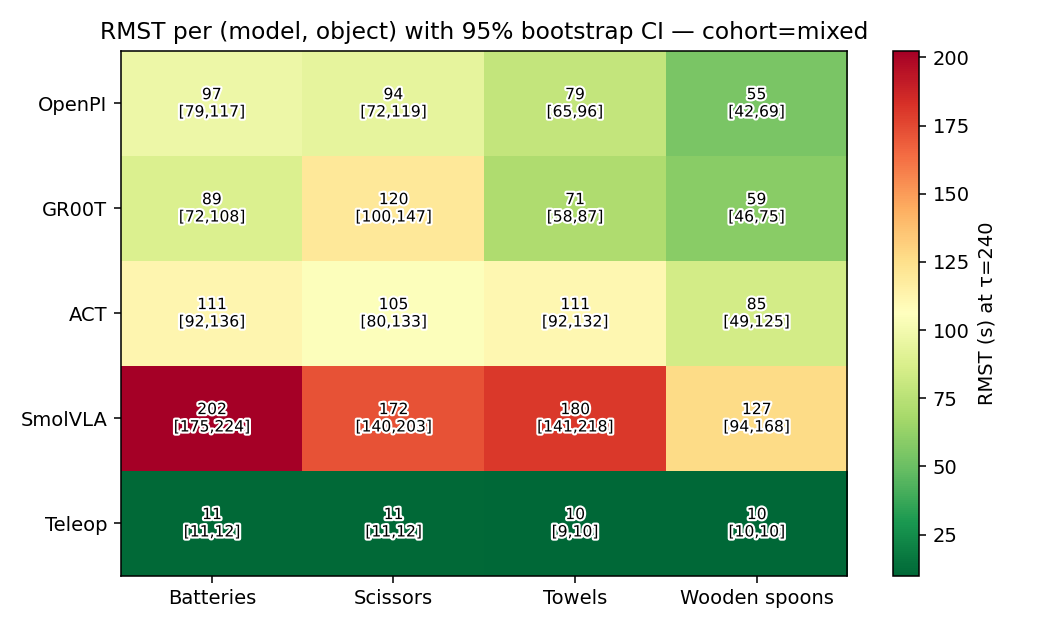}
  \caption{Per-(model, object) RMST with 95\% episode-clustered bootstrap CIs. Crossings exist (most clearly GR00T vs.\ ACT between Batteries and Scissors) but none clears Bonferroni at $\alpha = 0.05/24$. The visible patterns suggest the orderings vary by object; resolving them formally requires the rollout budget that Figure~\ref{fig:efficiency} quantifies.}
  \label{fig:grid-rmst}
\end{figure}

\begin{figure}[!h]
  \centering
  \includegraphics[width=0.92\linewidth]{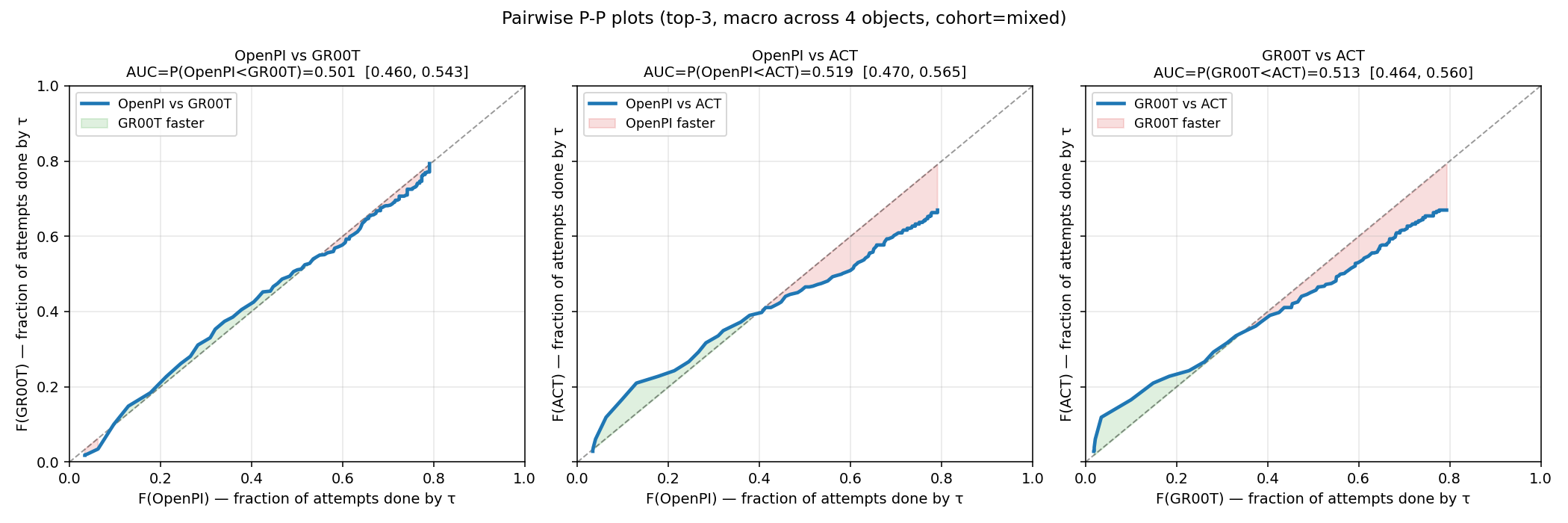}
  \caption{Pairwise model-vs-model P-P plots, top-3 only (SmolVLA excluded as too easily separable to be informative), macro-averaged across the 4 trained objects. The teleop-anchored AUC of \S\ref{sec:res-aggregation} saturates because the human reference is ${\sim}7\times$ faster than any evaluated VLA -- nearly all model events fall in the human's tail, so the AUC integral is dominated by short-time behaviour. Removing the human anchor neutralizes that bias: every top-3 pair falls on or near the diagonal, with macro-AUCs all consistent with $0.5$ (ACT-vs-OpenPI $0.51$\,[$0.45$, $0.56$]; ACT-vs-GR00T $0.50$\,[$0.45$, $0.55$]; OpenPI-vs-GR00T $0.50$\,[$0.46$, $0.54$]).}
  \label{fig:pp-pairwise}
\end{figure}

This appendix supports two claims from the body text. First, the per-(model, object) RMST grid (Figure~\ref{fig:grid-rmst}) substantiates \S\ref{sec:res-headline}'s 19\% per-object HRT ceiling and surfaces the GR00T-vs-ACT object crossing referenced in \S\ref{sec:res-aggregation}. Second, the pairwise P-P plots without the human anchor (Figure~\ref{fig:pp-pairwise}) accompany \S\ref{sec:res-aggregation}'s discussion of how the human-anchored AUC saturates and what removing that anchor reveals about the top-3 pairs.

\section{Distributional Shape and Scalar Trade-offs}\label{app:morevis}

The figures in this appendix complement the per-cell RMST grid in Appendix~\ref{app:percell} with two derived views of the same per-(model, object) CDFs: (i) a Q-Q view that exposes \emph{where} along the time axis the model lags the human reference, and (ii) a parametric trade-off view that shows two standard scalars (UPH and MTBF/A) pulling in opposite directions as the protocol's $\tau_\text{episode}$ varies.

\begin{figure}[!t]
  \centering
  \includegraphics[width=\textwidth]{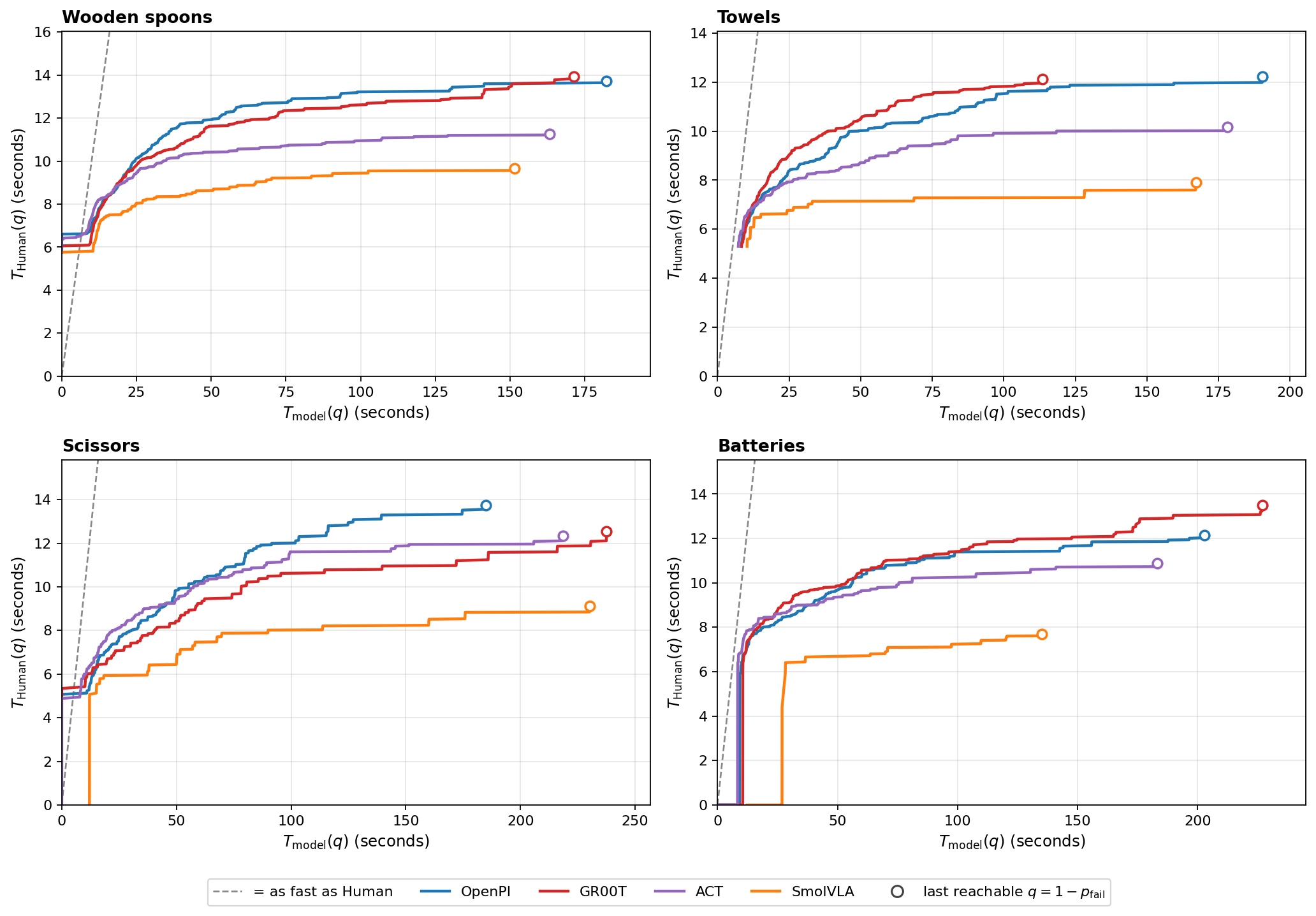}
  \caption{Per-object Q-Q plots, $T_\mathrm{Human}(q)$ vs $T_\mathrm{model}(q)$, all four VLAs on the four bin-to-bin objects. Dashed diagonal $=$ as fast as the human reference; below-diagonal $=$ model is slower. Open circle on each model curve marks $q=1-p_\mathrm{fail}$, the highest quantile the model reaches before its hard-failure asymptote. Curves bending below the diagonal as $q\to 1$ indicate tail-heavy slowdown rather than a uniform multiplicative gap.}
  \label{fig:qq-per-object}
\end{figure}

\paragraph{Q-Q plots: slowdown is tail-heavy.} Figure~\ref{fig:qq-per-object} plots, for each object, $T_\text{Human}(q)$ against $T_\text{model}(q)$ at matched quantile $q$ (i.e., the time by which a fraction $q$ of operations are completed). The dashed diagonal $T_\text{Human}=T_\text{model}$ marks ``as fast as the human reference''; below-diagonal points are model-slower-than-human. A uniform slowdown would trace a straight line through the origin with constant slope; the empirical curves bend away from the diagonal as $q\to 1$, indicating that policies fall behind disproportionately on the slow tail of operations rather than uniformly across the distribution. The open circle on each model curve marks the model's last reachable quantile $q=1-p_\text{fail}$: humans have effectively no hard-failure asymptote, so the human curve runs to $q\approx 1$ uninterrupted, while a model's curve terminates wherever its hard-failure rate cuts the CDF off.

\begin{figure}[!t]
  \centering
  \includegraphics[width=\textwidth]{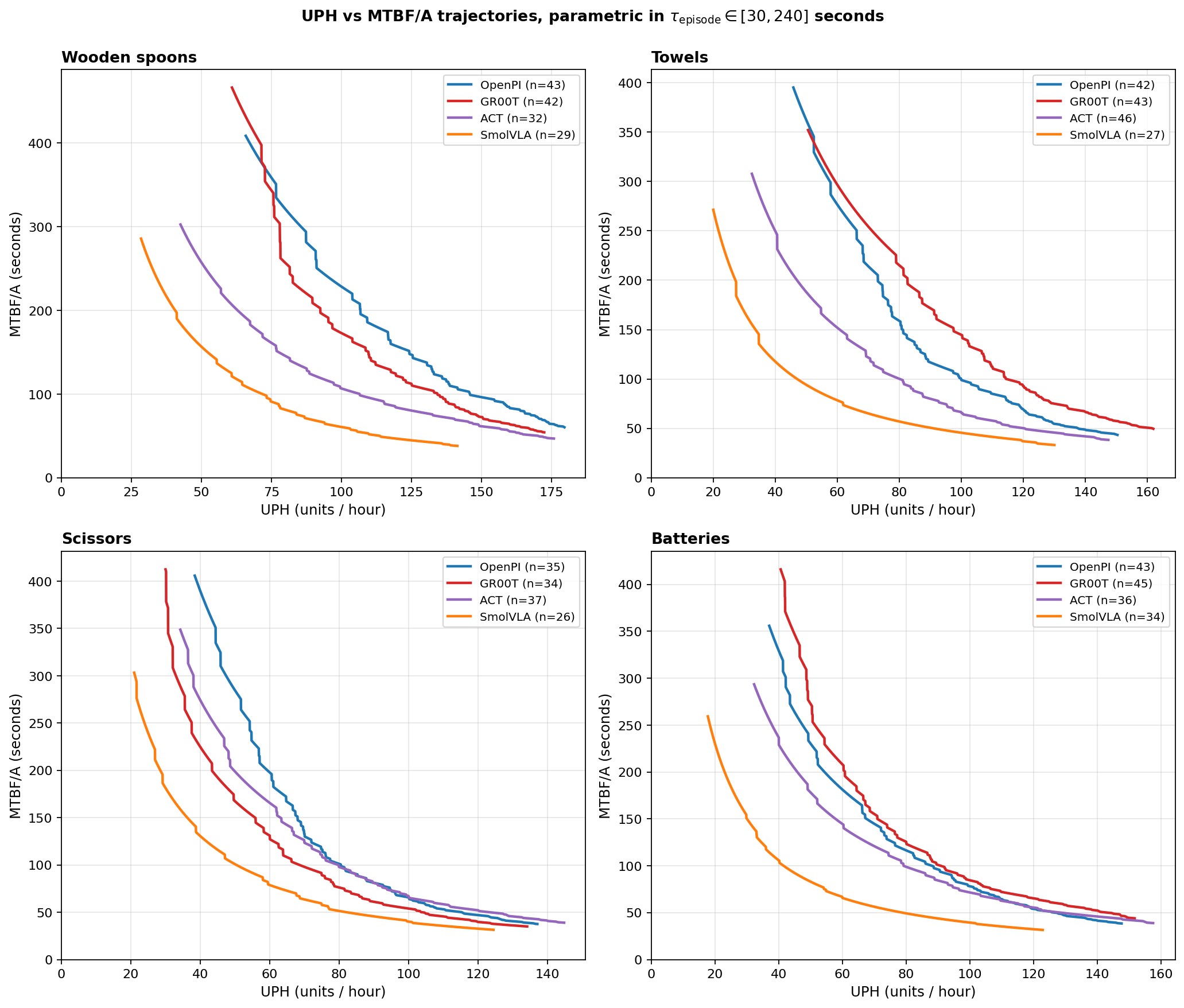}
  \caption{UPH versus MTBF/A trajectories per object, parametric in $\tau_\mathrm{episode}\in[30, 240]$ seconds. Each curve is one (model, object) cell traced as the episode horizon increases. Trajectories run up-and-left as $\tau_\mathrm{episode}$ rises (UPH falls, MTBF/A rises): the two scalars trade off, and the headline a benchmark publishes therefore depends on a protocol choice rather than only on policy quality.}
  \label{fig:uph-mtbf-per-object}
\end{figure}

\begin{figure}[!t]
  \centering
  \includegraphics[width=0.8\linewidth]{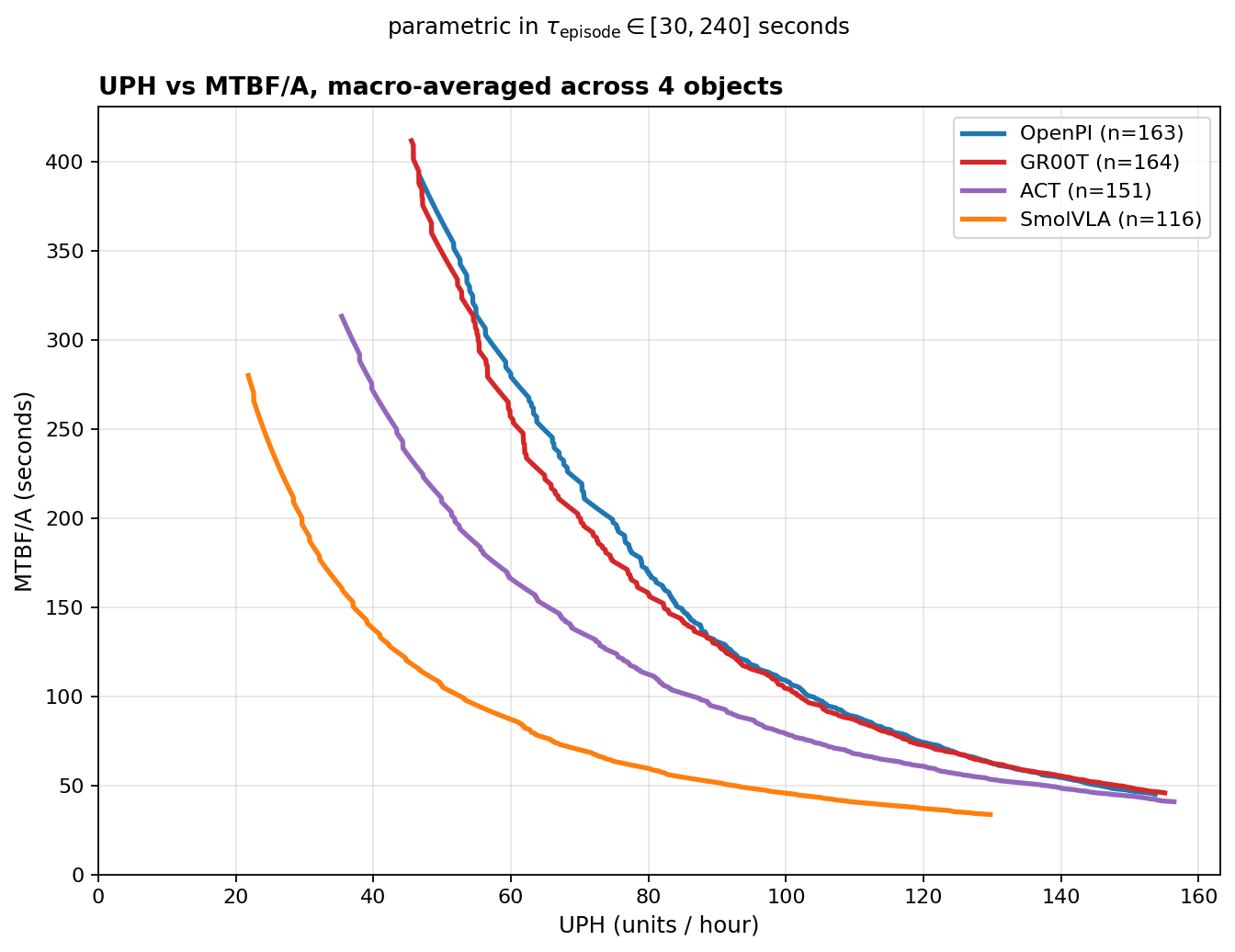}
  \caption{Macro-averaged UPH versus MTBF/A trajectories: for each model and each $\tau_\mathrm{episode}$, the four per-object (UPH, MTBF/A) points from Figure~\ref{fig:uph-mtbf-per-object} are averaged with equal weight across objects, then traced as $\tau_\mathrm{episode}$ sweeps $[30, 240]$ seconds. The UPH-vs-MTBF/A trade-off persists after macro-aggregation: any choice of $\tau_\mathrm{episode}$ fixes one point on each model's trade-off curve, and a different but equally defensible $\tau_\mathrm{episode}$ can change the headline ordering.}
  \label{fig:uph-mtbf-macro}
\end{figure}

\paragraph{UPH and MTBF/A trade off as $\tau_\mathrm{episode}$ varies.} Figure~\ref{fig:uph-mtbf-per-object} plots, for each object, every model's trajectory in the (UPH, MTBF/A) plane as $\tau_\mathrm{episode}$ sweeps $[30, 240]$ seconds. UPH $\propto 1/\mathrm{RMST}(\tau_\mathrm{episode})$ falls with $\tau_\mathrm{episode}$ because the integration cap charges every hard-failure ghost event the full $\tau_\mathrm{episode}$; MTBF/A $=\mathrm{RMST}(\tau_\mathrm{episode})/\big(1-F(\tau_\mathrm{episode})\big)$ rises with $\tau_\mathrm{episode}$ because the same cap inflates the numerator faster than the (eventually-asymptotic) denominator. The two scalars therefore move in opposite directions, and the order they induce on policies depends on which $\tau_\mathrm{episode}$ a paper picked. This is concrete evidence for the ``no single scalar suffices'' claim of \S\ref{sec:framework-scoring}: any headline scalar that integrates against $\tau_\mathrm{episode}$ implicitly fixes a point on a trade-off curve, and a different but equally defensible $\tau_\mathrm{episode}$ relabels the leaderboard. Figure~\ref{fig:uph-mtbf-macro} collapses the four per-object panels into a single macro view by equal-weight averaging the per-object (UPH, MTBF/A) trajectories at each $\tau_\mathrm{episode}$; the trade-off survives macro-aggregation, so it is a property of the methodology, not of a particular object.

\end{document}